\documentclass[MS]{macro/neu_msthesis}

\usepackage{physics}
\usepackage{siunitx}
\usepackage{graphics} % for pdf, bitmapped graphics files
\usepackage{graphicx} % for epx
\usepackage{epsfig} % for postscript graphics files
\usepackage{lipsum} 
\usepackage{amsmath, bm} % assumes amsmath package installed
\usepackage{amssymb}  % assumes amsmath package installed
\usepackage{upgreek}
\usepackage{url}
\usepackage{breakurl}
\usepackage{caption}
\usepackage{subcaption}
\usepackage{pgfplots}
\usepackage{floatrow}
\usepackage{float}
\usepackage[ruled,linesnumbered]{algorithm2e}
\usepackage{multicol}
\usepackage{multirow}
\usepackage{tikz}
\usepackage{booktabs}
\usepackage{longtable}
\usetikzlibrary{shapes, arrows}
\usepackage{amsmath}
\newfloatcommand{capbtabbox}{table}[][\FBwidth]

\title{Thrust Regulation Through Wing Linkage Modulation on the Aerobat
Platform: Piezoelectric Slip-Stick Actuated Regulator Development}

\author{Luca Ciampaglia}

\dept{Mechanical and Industrial Engineering}

\degree{Master of Science}

\degreename{Robotics}

\field{Robotics}

\submitdate{April 2026}

\numberofmembers{2}

\principaladviser{Dr. Alireza Ramezani}
\firstreader{Dr. Rifat Sipahi}
\chairman{Dr.  Rifat Sipahi}

\dean{Dr. Gregory D. Abowd}

\usepackage{amsfonts}			% special math characters
\usepackage{times}		% use Postscript fonts

\newcommand{\ifno}[1]{}

\usepackage{multirow}
\makeindex
\usepackage{makeidx}
\usepackage{acronym}

\usepackage{url}
\usepackage{graphicx}
\ifnum\pdfoutput>0
\usepackage[pdftex,%                    % hyper-references for ps2pdf
bookmarks=true,%                        % generate bookmarks ...
bookmarksnumbered=true,%                % ... with numbers
hypertexnames=false,%                   % needed for correct links to figures
breaklinks=true           %Allows link text to break across lines;
]{hyperref}
\else
\usepackage[hypertex,%                  % hyper-references for ps2pdf
bookmarks=true,%                        % generate bookmarks ...
bookmarksnumbered=true,%                % ... with numbers
hypertexnames=false,%                   % needed for correct links to figures
breaklinks=true           %Allows link text to break across lines;
]{hyperref}
\fi

\hypersetup{				% setup PDF information fields
pdfauthor = {\authorRef},
pdftitle = {\titleRef},
pdfsubject = {\expandafter{\degreeRef} thesis submitted to Northeastern University},
pdfkeywords = {Multimodal robots,M4,Traversability,3D path planning}
pdfcreator = {LaTeX with hyperref package},
pdfproducer = {dvips + ps2pdf}}

\begin{document}

% add a pdf bookmark to the cover page
\pdfbookmark[1]{Cover}{cover}

% --- title page ---
\titlepage

% --- front matter ---
\begin{frontmatter}
% print signature page
%\signaturepage
% dedication
%\input{tex/dedication.tex}

% table of content (add bookmark for convenience)
\pdfbookmark[1]{Table of Contents}{contents}
\tableofcontents
\listoffigures
\newpage\ssp
\listoftables

% include a list of Acronyms (comment out if no acronyms are specified)
% acronyms.tex

\chapter*{List of Acronyms}
\addcontentsline{toc}{chapter}{List of Acronyms}

\begin{acronym}
\acro{3D}{3-Dimensional}
\acro{ASME}{American Society of Mechanical Engineers}
\acro{ATMO}{Aerially Transforming Morphobot}
\acro{B2}{Bat Bot}
\acro{BLDC}{Brushless Direct Current}
\acro{CDC}{IEEE Conference on Decision and Control}
\acro{CEIM}{Contact-based Elastoplastic Independent Micro-contact}
\acro{CNC}{Computer Numerical Control}
\acro{DEA}{Dielectric Elastomer Actuator}
\acro{DOA}{Degree Of Actuation}
\acro{DOF}{Degree Of Freedom}
\acro{EPFL}{École Polytechnique Fédérale de Lausanne}
\acro{FDM}{Fused Deposition Modeling}
\acro{FWR}{Flapping Wing Robot}
\acro{GPS}{Global Positioning System}
\acro{ICRA}{IEEE International Conference on Robotics and Automation}
\acro{IEEE}{Institute of Electrical and Electronics Engineers}
\acro{MAV}{Micro Air Vehicle}
\acro{MIMIC}{Morphing via Integrated Mechanical Intelligence and Control}
\acro{PCA}{Principal Component Analysis}
\acro{PCB}{Printed Circuit Board}
\acro{PWM}{Pulse Width Modulation}
\acro{PZT}{Lead Zirconate Titanate}
\acro{RC}{Radio Controlled}
\acro{RSS}{Robotics: Science and Systems}
\acro{SIMP}{Solid Isotropic Material with Penalization}
\acro{SLA}{Stereo-Lithography Apparatus}
\acro{SMA}{Shape Memory Alloy}
\acro{TULA}{Tiny Ultrasonic Linear Actuator}
\acro{UWB}{Ultra-Wideband}
\end{acronym}

% include any of the front matter files that contain text
% attention the input does cause a page break, the include on 
% the other hand does not
% acknowledgements.tex:

\begin{acknowledgements}

I extend my heartfelt gratitude to all the members of the \textbf{SiliconSynapse Lab} and Dr. Alireza Ramezani for his mentorship and advice.

\end{acknowledgements}

% abstract.tex:

\begin{abstract}

Aerobat is a bat-inspired flapping-wing robot with a wing gait generate by the computational structure, a planar linkage of carbon fiber links driven by a single motor. This design minimizes weight but couples both wings to a shared input motor, eliminating independent thrust control and preventing asymmetric maneuvers. This thesis investigates thrust regulation by modifying the effective length of the first radius link $R_1$ in the computational structure. Static experiments using FDM-printed $R_1$ links at three lengths (28.58, 29.33, and 30.08~mm) across 3,4, and 5~Hz flapping frequencies demonstrated that a 1.5~mm length increase produced a 37\% increase in peak lift force and shifted peak force timing within the downstroke. An additional experiment using a string-actuated regulator mechanism was performed. Further actuation methods were evaluated: sub-gram micro-servo and piezoelectric slip-stick. After both the string-tension and micro-servo actuation methods failed due to structural member compliance and motor fragility respectively, a TULA-50 piezoelectric slip-stick actuator was selected. Multiple force-amplifying mechanisms were prototyped, resulting in a direct-drive variable-length mechanism. This final mechanism was demonstrated in a preliminary bench-top test, though insufficient force output prevented dynamic testing during flapping. This work establishes linkage-length modulation via embedded slip-stick actuation as a viable approach to independent wing thrust control.

\end{abstract}

\end{frontmatter}

% --- body of the document ---

%\pagestyle{plain}
\pagestyle{headings}

% include each chapter like below
\chapter{Introduction}
\label{chap:Introduction}

\section{Background and Motivation}
Biological flyers such as bats possess flight capabilities that no traditional thruster-actuated aerial platform can replicate. Bats navigate cluttered environments, execute rapid banking turns, perch on irregular surfaces, and transition between flight gaits, all within a body mass of less than 100 grams~\cite{hedenstrom2015}. These behaviors are enabled by a musculoskeletal wing architecture comprising more than 40 kinematic degrees of freedom~\cite{swartz2008}, which allows the animal to continuously reshape its wing membrane throughout each flap cycle. Unlike fixed-wing aircraft, which require sustained forward velocity, or rotary-wing drones, whose continuous rotor downwash destabilizes flight near surfaces, flapping-wing robots \ac{FWR}s generate aerodynamic forces through oscillatory wing motion that can be dynamically shaped within each stroke~\cite{nekoo2025}. These capabilities arise not merely from flapping, but from the ability to produce asymmetric and independently modulated forces across the left and right wings, enabling the rapid banking turns, pitch corrections, and heading changes that define agile biological flight~\cite{iriartediaz2008}. Replicating this performance in a robotic system demands an actuation architecture that can differentially control the thrust produced by each wing, a challenge that remains largely unsolved at the small scales where morphing-wing flight is most advantageous~\cite{phan2019}.

\begin{figure}[ht]
    \centering
    \includegraphics[height=0.5\linewidth]{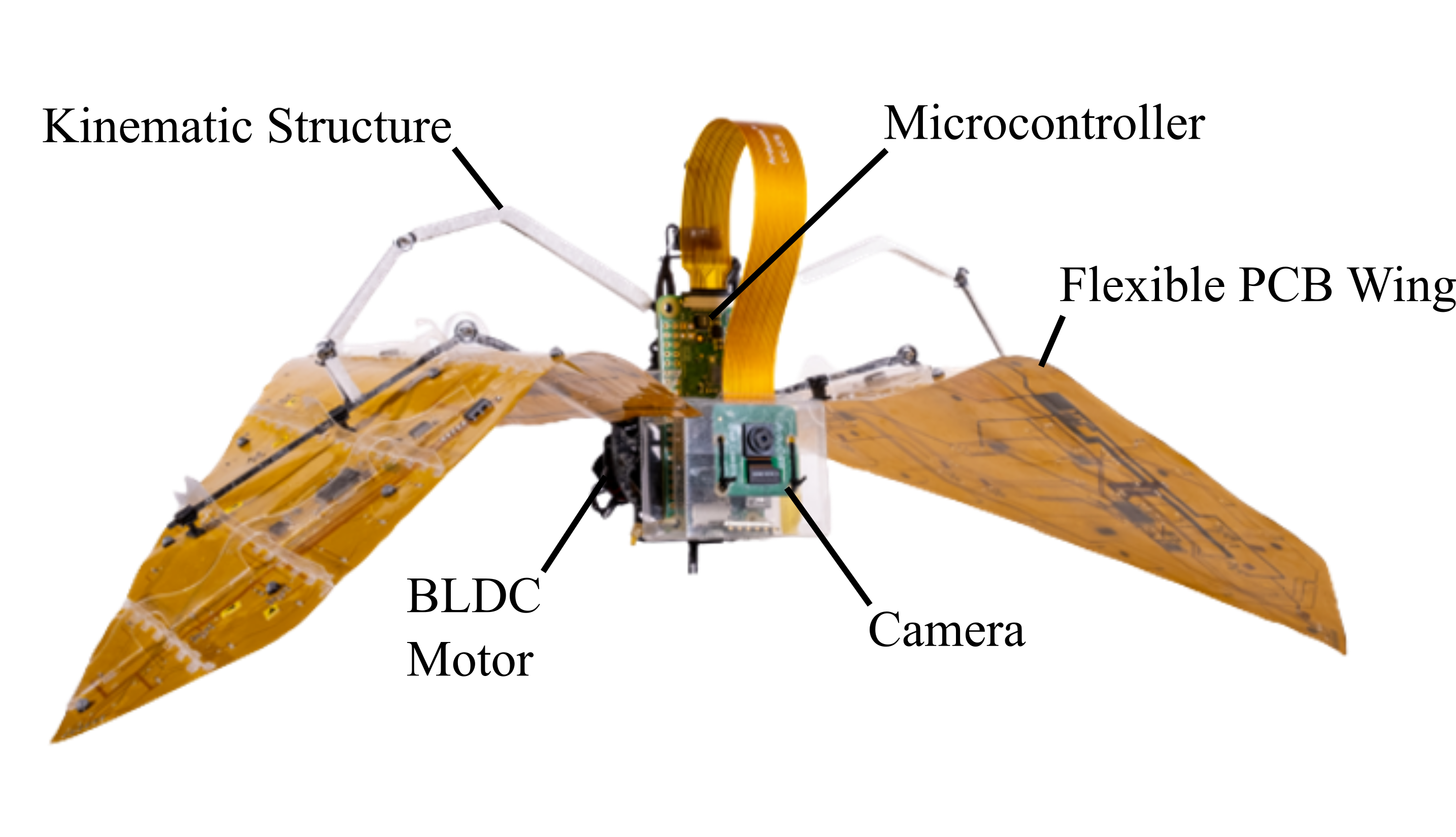}
    \caption{The Aerobat platform. The kinematic wing structure, composed of rigid carbon fiber links and revolute joints, generates the flapping gait from a single motor input. Image adapted from~\cite{gupta2025forces}}
    \label{fig:aerobat_intro}
\end{figure}

Aerobat is a bat-inspired \ac{FWR} developed in the Silicon Synapse Lab at Northeastern University, designed to study dynamic morphing wing flight in confined environments~\cite{sihite2024ijrr, ramezani2022aerobat}. The core of Aerobat's mechanical design is a computational structure: a planar linkage of rigid carbon fiber links connected by revolute joints, driven by a single \ac{BLDC} motor through a gear train. Rather than requiring a dedicated actuator for each joint in the kinematic chain, the geometry of this linkage is optimized such that rotation of the motor shaft produces a cyclical flapping gait consisting of wing plunging, elbow flexion-extension, and membrane morphing through the kinematic constraints of the mechanism itself~\cite{sihite2020armwing}. This design philosophy aligns with the Morphing via Integrated Mechanical Intelligence and Control (MIMIC) framework~\cite{sihite2021mimic}, which achieves flight control through small, low-power primer actuators that modify the computational structure's geometry rather than relying solely on closed-loop feedback. This approach dramatically reduces actuator count and weight, making it well suited to the extreme mass constraints of a sub-25 gram aerial platform. Figure~\ref{fig:aerobat_intro} shows the Aerobat Gamma platform. The current iteration, Aerobat Delta, features a single wing and is used as the primary test platform in this thesis.

The computational structure that allows Aerobat to be driven by a single motor also introduces a fundamental control limitation. Because both wings are coupled to a single motor through a shared gear train, there is no means of independently modulating the thrust produced by each wing. Increasing or decreasing motor speed affects both wings symmetrically. This provides only a single scalar control input, motor RPM, that governs the flapping frequency of both wings identically. This eliminates roll authority, removes any mechanism for heading control, and prevents the execution of any maneuver requiring asymmetric aerodynamic forces. Prior work from the lab has addressed this limitation through a guard system with manifold small thrusters~\cite{dhole2023guard}, which stabilizes Aerobat's position and orientation during flight but adds significant mass and negates the aerodynamic advantages of morphing-wing design. However, this was never intended to be a permanent solution. The implementation of a necessary degree of actuation for the purposes of aerial roll authority is the challenge that this thesis addresses.

This problem is further compounded by the mass constraints of the platform. The current Aerobat prototype has a measured mass of approximately 25~g, while the wings produce approximately 35~g of thrust, leaving a payload margin of roughly 10~g for all additional onboard systems including electronics, sensing, and any control mechanisms. This narrow margin means that any mechanism added to provide thrust regulation must be designed to contribute minimal mass, require few components, and employ an actuation technology with a sufficiently high power-to-weight ratio for the given scale~\cite{karpelson2008, helbling2018}. Adding a conventional servo-based or motor-driven actuation system, similar to what was used at larger scales~\cite{gerdes2014roboraven, huang2022ustbird}, would likely consume most or all of the remaining payload budget, leaving no capacity for other essential subsystems. 

\section{Approach and Contributions}
The driving concept of this thesis is that modifying the effective length of the first radius link in the computational structure alters the wing's flapping gait envelope and thereby modulates the thrust produced by a single wing. Because this link connects the motor-driven shaft to the rest of the wing linkage, even sub-millimeter changes in its length shift the kinematic trajectory of the elbow joint and wingtip throughout the flapping cycle, producing measurable differences in both peak aerodynamic force and the temporal distribution of that force within each stroke. This approach is consistent with the MIMIC framework's principle of embedding small geometric modifications into the computational structure to achieve control authority~\cite{sihite2021mimic}. This hypothesis was first validated experimentally using a set of \ac{FDM}-printed radius links of discrete lengths, tested on the Aerobat Delta prototype mounted to a 6-axis load cell. With the hypothesis confirmed, multiple methods of actuation were evaluated for use in controlling the length of the first radius link. After several iterations, a mechanism was designed and fabricated using a piezoelectric slip-stick actuator to achieve this length change dynamically, replacing the rigid first radius link with a variable-length prismatic joint embedded directly in the wing structure. The selection of this actuator followed an iterative process in which string-tension and sub-gram micro-servo approaches were first evaluated and found to be inadequate for the platform's weight and performance requirements.
 
The specific contributions of this thesis are as follows:
\begin{enumerate}
    \item Experimental validation that variation in the first radius link length produces significant changes in thrust magnitude and flapping gait timing, established through static testing with fixed-length printed links at multiple flapping frequencies.
    \item Systematic evaluation of several candidate actuation methods including string-tension, sub-gram microservo, and piezoelectric slip-stick approaches, against the weight, stroke, force, and integration constraints of the Aerobat platform, with documented failure modes and design lessons from each iteration.
    \item Design and fabrication of a piezoelectric slip-stick actuated prismatic joint mechanism achieving the required 1.5~mm stroke within the geometry of the first radius link.
    \item Benchtop demonstration of the mechanism's ability to modulate linkage length, with mechanical integration into the Aerobat Delta airframe. 
\end{enumerate}

\section{Thesis Organization}
The remainder of this thesis is organized as follows. Chapter~\ref{chap:lit-review} reviews the relevant literature across seven areas: piezoelectric slip-stick actuator principles, piezoelectric actuation in flapping-wing robots, mechanical design and transmission mechanisms for \ac{FWR}s, slip-stick actuation in micro-robotics, comparative actuator technology studies, servo and cable-tendon based wing control, and prior work from the Silicon Synapse Lab on bat-inspired morphing-wing platforms. The chapter concludes by identifying the specific research gap this thesis addresses. Chapter~\ref{chap:Methodology} describes the Aerobat Delta platform, the static \ac{FDM}-printed regulator experiments, the derivation of performance requirements, and the actuator selection and iterative design process leading to the final piezoelectric regulator mechanism. Chapter~\ref{chap:results} presents the experimental results from both the static link length tests and the rudimentary benchtop mechanism validation. Chapter~\ref{chap:conclusion} discusses the findings, limitations of the current work, and directions for future research including dynamic flight testing and closed-loop control integration.

% Path planning

\chapter{Literature Review}
\label{chap:lit-review}
This chapter reviews the literature relevant to the design and development of a thrust-regulating mechanism for the Aerobat flapping-wing robot. The review is organized into four sections. Section~\ref{sec:fwr_design} surveys flapping-wing robot mechanical design, covering linkage-based actuation mechanisms, servo and cable-tendon approaches to wing joint control, wing morphing and variable geometry strategies, and comparative actuator technology studies. Section~\ref{sec:piezo} focuses on piezoelectric actuation, reviewing its use as a primary flight actuator in flapping-wing MAVs and the principles, modeling, and design of slip-stick piezoelectric actuators specifically. Section~\ref{sec:ssl} summarizes prior work from the Silicon Synapse Lab on bat-inspired morphing-wing platforms, culminating in the Aerobat system that motivates the present work. Section~\ref{sec:gap} identifies the specific research gap this thesis addresses.\

\section{Flapping Wing Robot Design and Actuation}
\label{sec:fwr_design}
The design of flapping-wing robot actuation mechanisms is fundamentally informed by the biomechanics of biological flyers. Among flying vertebrates, bats possess the most kinematically complex wings, with over forty degrees of freedom arising from elongated forelimb bones, multiple digit joints, and a compliant membrane spanning the digits, body, and hindlimbs~\cite{swartz2008, hedenstrom2015}. This skeletal architecture allows bats to continuously reshape their wing geometry throughout each flap cycle, modulating camber, angle of attack, wing area, and stroke plane on a sub-wingbeat timescale~\cite{riskin2010}. Critically, bats achieve maneuvering flight through asymmetric modulation of left and right wing kinematics: during turning, the inner wing exhibits reduced stroke plane angle and flaps closer to the body, while the outer wing sweeps through a steeper, wider arc, generating the differential aerodynamic forces required for roll and yaw~\cite{iriartediaz2008}. This biological strategy of achieving control authority through geometric modulation of individual wing parameters, rather than through separate control surfaces, directly motivates the approach taken in this thesis, where modulating a single linkage parameter in the wing's computational structure is used to differentially regulate thrust.

\subsection{Linkage-Based Actuation Mechanisms}
The mechanical integration of actuators into wing structures requires careful consideration of transmission design. Zhang and Rossi~\cite{zhang2017} provide a foundational review of compliant transmission mechanisms for \ac{FWR}s, covering how compliance reduces part count, weight, and friction while enabling elastic energy storage and release during the flapping cycle. Their optimization method, inspired by insect thoraxes, is particularly relevant since piezoelectric stick-slip actuators would likely require compliant coupling to the wing.

At bird scale, Ryu et al.~\cite{ryu2020} present two four-bar linkage wing mechanisms for combined flapping and folding, deriving kinematic relationships and optimizing link lengths, with experimental validation showing that incorporating folding improves both frequency and lift. Singh et al.~\cite{singh2022} provide the most detailed taxonomy available for mapping linkage topology to vehicle scale and flight capability, using novel structural block diagrams to classify flapping actuation mechanisms from over two decades of literature. Han et al.~\cite{han2023} focus specifically on avian-scale mechanisms, classifying them into one-axis and multi-axis categories, and critically note that compliant mechanisms become less suitable at avian scales due to lower transmission efficiency under larger wing loads.

The most recent and comprehensive review by Nekoo et al.~\cite{nekoo2025}, published in the \textit{International Journal of Robotics Research}, covers transmission mechanisms, wing design with functional group joints, actuator scaling laws, and the distinction between large-scale motor-driven systems and small-scale piezoelectric/\ac{DEA} systems. This review identifies open research challenges in wing morphing and perching and provides the systems-level perspective needed to situate the present thesis within the current state of the field.

A common thread across all of these works is that the linkage geometry is treated as a fixed design parameter: link lengths, joint positions, and transmission ratios are optimized during the design phase and remain constant during operation. Once fabricated, the mechanism produces a single flapping gait determined entirely by the input motor speed. No existing linkage-based FWR mechanism provides the ability to modify individual link lengths in-flight to alter the resulting gait and force output. This limitation is precisely what the present thesis seeks to address by introducing a variable-length element into the Aerobat's computational wing structure.

\subsection{Wing Morphing and Variable Geometry}
A separate body of work has explored modifying wing geometry during flight to alter aerodynamic performance, drawing inspiration from the shape-changing capabilities of biological wings. Colorado et al.~\cite{colorado2012} developed a bat-inspired robot using NiTi shape memory alloy wires as artificial biceps and triceps muscles to morph wing shape at the elbow joint, demonstrating measurable changes in aerodynamic forces through controlled wing folding and extension during the flap cycle. Di Luca et al.~\cite{diluca2017} designed a morphing wing composed of artificial feathers that fold and extend to change wing surface area, achieving up to $48\%$ drag reduction in the folded configuration and demonstrating asymmetric folding for roll control. At larger scales, Ajanic et al.~\cite{ajanic2023} combined wing folding, pitching, and feather extension via servo motors and cable-tendon transmission, showing through wind tunnel testing that wing morphing improves aerodynamic efficiency. Chen et al.~\cite{chen2023morphing} introduced an actuation strategy that couples morphing with the wingbeat cycle via servo-controlled linkages, producing larger control moments than conventional approaches.
These morphing approaches share a common strategy: they modify the wing's aerodynamic shape, whether camber, area, sweep, or twist, to change the forces generated. The present thesis takes a fundamentally different approach. Rather than altering the wing surface geometry directly, it modulates a structural parameter of the kinematic linkage mechanism itself, specifically the effective length of the first radius link, which in turn changes the entire flapping gait envelope. This distinction is important because modifying the linkage parameter affects not only the peak force magnitude but also the temporal force distribution within each stroke cycle, as demonstrated by the static experiments described in Chapter~\ref{chap:Methodology} and Chapter~\ref{chap:results}.

\subsection{Servo and Cable-Tendon Based Actuation}
An alternative approach to wing actuation uses servo motors, either directly driving wing joints or transmitting forces through cable-tendon systems. This strategy enables programmable, independent control of individual wing degrees of freedom, in contrast to the coupled kinematics of single-motor gear-driven systems.

The Robo Raven project, reported by Gerdes et al.~\cite{gerdes2014roboraven}, was the first demonstration of a bird-inspired platform performing outdoor aerobatics using independently actuated and controlled wings. Each wing is powered by its own programmable digital servo motor, enabling arbitrary motion profiles for exploring wing kinematics--force relationships. The retrospective by Bruck et al.~\cite{bruck2023roboraven} documents five successive Robo Raven designs, progressing from basic independent wing control through solar energy harvesting, autonomous decision-making, and mixed-mode propulsion. Huang et al.~\cite{huang2022ustbird} developed USTBird, an all servo-driven bird-like robot where two independently programmable servos directly drive each wing without gear trains, achieving decoupled three-axis control torque and agile aerobatic maneuvers without a controllable tail. A subsequent version, USTBird-I~\cite{huang2022ustbird1}, incorporated a camber wing structure and dihedral angle adjustment mechanism, demonstrating the first outdoor autonomous airdrop mission by a flapping wing robot.

Cable-tendon actuation provides an alternative force transmission path that can reduce weight at wing joints. Olejnik et al.~\cite{olejnik2018sweep} demonstrated a servo-and-string mechanism on a DelFly-based \ac{FWR}, where servo forces are transferred through strings to bend wing leading edges, modulating sweep angle independently on each of four wings for roll and pitch control. Ajanic et al.~\cite{ajanic2023} developed a highly articulated biohybrid robotic wing with three servo motors controlling flapping, pitching, and folding via a differential gear and tendon-based extension/retraction mechanism using real bird feathers. Wind tunnel testing revealed that wing folding and stroke tilting improve aerodynamic efficiency. Bahlman et al.~\cite{bahlman2013} constructed a multi-articulated robotic bat wing modeled after \textit{Cynopterus brachyotis} with seven joints powered by three servo motors, enabling simultaneous measurement of power input and force output across a range of kinematic parameters.

At larger scales, recent work has achieved increasingly sophisticated servo-driven wing control. Wang et al.~\cite{chen2025robofalcon} developed RoboFalcon2.0, an 800~g, 1.2~m wingspan robot with reconfigurable mechanisms where high-voltage servos control folding and sweeping amplitude, achieving the first bird-style self-takeoff in a robot, as reported in \textit{Science Advances}. Liu et al.~\cite{liu2025hithawk} designed HIT-Hawk with coordinated wing-tail distance adjustment using servo-driven morphing, achieving 13 Dynamic Flying Primitives. Chen et al.~\cite{chen2023morphing} introduced a novel actuation strategy coupling morphing with the wingbeat cycle via servo-controlled linkages, demonstrating larger control moments than conventional approaches in \textit{\ac{IEEE} Transactions on Robotics}. The Festo BionicSwift~\cite{festo2020bionicswift} demonstrated an ultralight 42~g robotic bird with overlapping foam wing segments controlled by two servos, achieving coordinated multi-robot flight using \ac{UWB} indoor \ac{GPS}. Karasek et al.~\cite{karasek2014pitch} presented a servo-driven mechanism modulating flapping amplitude and offset for pitch and roll control of a hovering \ac{FWR}, identifying that the controlled mechanism consumes up to twice the power of an uncontrolled prototype.

While servo and cable-tendon approaches have proven effective for achieving independent wing control, they carry inherent weight and complexity costs that become increasingly prohibitive at smaller scales. The platforms described above operate at masses ranging from 42~g (BionicSwift) to 800~g (RoboFalcon2.0), with most in the 100 to 500~g range. Even the lightest servo-driven systems use servos for whole-wing actuation such as flapping, folding, or sweep, not for modulating individual link geometry within a kinematic chain. At the Aerobat's operating scale of approximately 25~g with a payload margin of roughly 10~g, integrating even a sub-gram servo for each wing's link regulation consumes a significant fraction of the available mass budget, particularly when accounting for mounting hardware, wiring, and structural reinforcement. This scale mismatch motivated the exploration of the alternative actuation technologies of micro-servo and piezoelectric slip-stick actuation, as described in Chapter~\ref{chap:Methodology}.

\subsection{Comparative Actuator Technologies}
Systematic actuator selection is critical for \ac{FWR} design. Karpelson et al.~\cite{karpelson2008} published the foundational actuator selection paper, evaluating piezoelectric, electromagnetic, \ac{SMA}, thermal, and electrostatic technologies against mechanical requirements and identifying piezoelectric actuation as most promising based on power density and bandwidth at insect scales. Helbling and Wood~\cite{helbling2018} updated this analysis with a decade of progress, reporting piezoelectric power density of approximately 467~W/kg for the RoboBee platform and comparing against electromagnetic motors with approximately 60\% gear-train efficiency. Their finding that onboard power storage---not actuation---is the most significant bottleneck for autonomous flight is essential for understanding system-level constraints.

Chen and Zhang~\cite{chen2019} review \ac{FWR}s organized by actuation method, emphasizing that power electronic topologies must be co-designed with actuator selection, as stick-slip drives that require asymmetric waveform generation require complex driving circuits. Phan and Park~\cite{phan2019} provide a cross-platform comparison of vehicles achieving free flight, demonstrating how actuator choice fundamentally constrains vehicle scale, endurance, and control authority. Hassanalian and Abdelkefi~\cite{hassanalian2019} evaluate conventional flapping mechanisms at the bird scale ($\sim$100~g), presenting a novel hybrid six-bar mechanism optimized for upstroke/downstroke asymmetry. McIvor and Chahl~\cite{mcivor2020} offer one of the few quantitative comparisons between electromagnetic and piezoelectric actuators, showing that resonance tuning reduces electromagnetic power consumption to 13\% of baseline while reporting piezoelectric power density of approximately 467~W/kg. Xiao et al.~\cite{xiao2021} classify hoverable \ac{FWR}s into motor-driven and intelligent-actuator-driven categories with tabulated side-by-side performance data useful for quantitative benchmarking.

Across these comparative studies, piezoelectric actuation consistently emerges as the technology with the highest power density and bandwidth at small scales, while electromagnetic motors dominate at bird and larger scales due to their higher absolute force output and mature control infrastructure. However, all of these comparisons evaluate actuators as primary flight drivers, meaning the motor or piezoelectric element that generates the flapping motion itself. The use case in this thesis is different: the actuator serves as a secondary regulator that modulates a linkage parameter within an existing motor-driven structure. For this application, the relevant performance metrics shift from raw power output to resolution, compactness, holding force, and mass. These criteria favor piezoelectric slip-stick actuators over both electromagnetic motors and piezoelectric bending actuators, as discussed in the following section.

\section{Piezoelectric Actuation for Flapping Flight and Micro-Robotics}
\label{sec:piezo}
Piezoelectric materials convert electrical energy to mechanical displacement with high bandwidth, high power density, and minimal moving parts, making them attractive for actuation at small scales. This section reviews two distinct applications of piezoelectric technology relevant to this thesis. The first is the use of piezoelectric bending actuators (bimorphs and unimorphs) as primary flight actuators in insect-scale FWRs. The second is the family of piezoelectric slip-stick (inertial) actuators, which produce linear translation through asymmetric friction-inertia cycles and have been extensively developed for precision positioning in micro-robotics but have not previously been applied to flapping-wing systems. Understanding both bodies of work is necessary to appreciate why slip-stick actuation was selected for the Aerobat regulator mechanism despite the dominance of bending actuators in existing piezoelectric \ac{FWR} literature.

\subsection{Piezoelectric Bending Actuators in Flapping Wing MAVs}
The landmark demonstration of piezoelectric actuation for flapping flight was achieved by Ma et al.~\cite{ma2013}, whose 80~mg RoboBee used piezoelectric bimorph actuators with a flexure-based four-bar transmission to achieve the first controlled flight of an insect-scale robot at approximately 120~Hz flapping frequency. This work, published in \textit{Science}, established \ac{PZT} bimorphs as the dominant actuator technology for sub-gram \ac{FWR}s. Ozaki and Hamaguchi~\cite{ozaki2018} subsequently introduced a direct-drive piezoelectric approach that eliminates separate transmission mechanisms entirely, demonstrating takeoff with a 114~mm wingspan prototype. This simpler architecture reduces part count and structural complexity, offering an alternative integration strategy.

Further milestones in energy autonomy followed rapidly. Jafferis et al.~\cite{jafferis2019} reported the first sustained untethered flight of the 259~mg RoboBee X-Wing in \textit{Nature}, achieving power consumption of only 110--120~mW with integrated solar cells. Yang et al.~\cite{yang2019} demonstrated Bee+, a 95~mg four-winged robot using 28~mg twinned piezoelectric unimorph actuators with four independently actuated wings providing full attitude control including yaw. Ozaki et al.~\cite{ozaki2021} demonstrated RF wirelessly powered flight of a 1.8~g \ac{FWR} with a power-to-weight density of 4,900~W/kg, published in \textit{Nature Electronics}. The current state of the art in piezoelectric \ac{FWR} energy autonomy is represented by the 2.1~g battery-powered untethered vehicle of Ozaki et al.~\cite{ozaki2023}, which achieved over five minutes of operation at 1.5$\times$ body weight in thrust using direct-drive piezoelectric unimorph actuators with passive charge recovery.

All of the piezoelectric FWR systems described above use bending-mode actuators, whether bimorphs or unimorphs, as the primary source of flapping motion. The piezoelectric element itself oscillates at the flapping frequency, and its resonant behavior is integral to achieving sufficient wing stroke amplitude and energy efficiency. In these architectures, the actuator \textit{is} the flight motor. The present thesis employs piezoelectric actuation in a fundamentally different role. Rather than driving the flapping motion, the piezoelectric actuator serves as a secondary regulator embedded within a motor-driven computational structure, making quasi-static adjustments to a linkage parameter between or during flap cycles. This use case does not require resonant oscillation or high-frequency cycling. Instead, it demands precise linear displacement, adequate force output under load, compact form factor, and the ability to hold a commanded position without continuous power input. These requirements align not with bending-mode actuators but with a different class of piezoelectric device: the slip-stick inertial actuator.

\subsection{Piezoelectric Slip-Stick Actuators: Principles, Design, and Optimization}

The concept of piezoelectric inertial sliding was first introduced by Pohl~\cite{pohl1987}, who demonstrated that a sawtooth-driven piezoelectric element produces stepwise translation via slow extension (stick) followed by rapid retraction (slip), achieving step sizes of 0.04--0.2~$\mu$m and speeds up to 0.2~mm/s under 1~kg loads. This foundational work established the operating principle upon which all subsequent stick-slip actuator research is built. Renner et al.~\cite{renner1990} extended the concept to vertical operation against gravity for scanning tunneling microscope approach mechanisms, proving that inertial drive forces can overcome gravitational loading---a key consideration for actuators integrated into wing structures that experience varying orientations during flight.

The most widely cited overview of the field is Hunstig's comprehensive review~\cite{hunstig2017}, which systematically covers the historical development, four distinct operating modes (two stick-slip and two slip-slip variants), friction contact design, waveform optimization, and control strategies. This review resolves the widespread terminological confusion between ``stick-slip,'' ``inertial,'' and ``impact'' drive nomenclature. Li et al.~\cite{li2019stepping} compare the three main stepping piezoelectric actuator families---inchworm, friction-inertia (stick-slip), and parasitic types---providing a systematic framework for justifying stick-slip selection over competing piezoelectric approaches based on resolution, speed, stroke, and load capacity metrics.

More recent work has focused specifically on mechanical design. Qiao et al.~\cite{qiao2022} review stick-slip actuators incorporating flexure hinge mechanisms, categorizing designs into translatory, rotary, and multi-\ac{DOF} types and comparing performance across lever, triangular, parallelogram, and rhombus flexure topologies. This work directly addresses the structural integration challenge central to the present thesis. Lin et al.~\cite{lin2024} provide a 45-page survey covering longitudinal and shear driving modes, compliant mechanism topology optimization including \ac{SIMP} methods, and performance benchmarking across diverse configurations. The most recent comprehensive review by Zhong et al.~\cite{zhong2025} covers driving signal optimization, structural design innovations, and friction modeling approaches including LuGre, elastoplastic, and neural-network-based methods. This review identifies open challenges in miniaturization and friction management that are directly relevant to scaling stick-slip actuators for integration into \ac{FWR}s.

Several properties of slip-stick actuators make them particularly attractive for the Aerobat regulator application. First, they offer sub-micron stepping resolution, which provides fine control over the linkage length adjustment well within the 1.5~mm target stroke. Second, the stroke is theoretically unlimited since displacement accumulates over successive steps, meaning that the actuator's travel is constrained by the mechanical design of the housing rather than by the piezoelectric element's intrinsic strain. Third, slip-stick actuators hold their position passively through static friction when the drive signal is removed, eliminating the need for continuous power to maintain a commanded length. Finally, their compact, solid-state construction with no gears, bearings, or rotating components makes them well suited to integration within the tight geometric envelope of the first radius link. These characteristics collectively distinguish slip-stick actuators from the other actuation technologies evaluated in this thesis, including string-tension drives, microservos, and piezoelectric bending actuators.

\subsection{Slip-Stick Actuation Principles in Micro-Robotics}

The theoretical and practical foundations of stick-slip drives in micro-robotics extend beyond \ac{FWR}s. Breguet and Clavel~\cite{breguet1998} demonstrated stick-slip actuators in several configurations at \ac{EPFL}, including a 1-\ac{DOF} actuator, a 6-\ac{DOF} platform, and two 3-\ac{DOF} mobile microrobots, achieving centimeter-range displacements at mm/s speeds with sub-5~nm resolution using monolithic piezoceramic flexible structures. This work established one of the most influential research groups in the field.

Detailed modeling efforts have been critical for understanding actuator dynamics. Edeler et al.~\cite{edeler2011} introduced the \ac{CEIM} friction model and validated it against experimental data for miniaturized handling robots. Peng and Chen~\cite{peng2011}, publishing in \textit{\ac{ASME} Transactions on Mechatronics}, integrated piezoelectric hysteresis modeling with elastoplastic friction models to capture the full nonlinear dynamics, providing a foundation for both design optimization and closed-loop control necessary for precise, repeatable wing actuation. Zhang et al.~\cite{zhang2012} generalized various friction--inertia actuator designs into a unified framework with three operating principles, identifying miniaturization limits and environmental sensitivity as open issues relevant to the challenging operating conditions of a flapping wing robot.

These modeling and experimental efforts in the micro-robotics community provide the theoretical foundation upon which the actuator selection and mechanism design in this thesis are built. The friction dynamics characterized by Edeler, Peng, and Zhang govern the force transmission and stepping behavior of the TULA-50 actuator used in the Aerobat regulator. Furthermore, the miniaturization challenges and environmental sensitivity identified by Zhang et al. are directly relevant to the operating conditions of a flapping wing robot, where the actuator must perform reliably under cyclic vibrations, varying orientations, and aerodynamic loading. The successful demonstration of slip-stick drives at the scales and resolutions reported by Breguet and Clavel confirms that the technology is viable at the size and mass constraints of the Aerobat platform.

\section{Previous Work}
\label{sec:ssl}

The thrust regulation mechanism developed in this thesis builds directly on the platforms and design philosophy established within the Silicon Synapse Lab at Northeastern University. This section traces the evolution from the foundational Bat Bot platform through the current Aerobat system, with particular emphasis on the computational structure concept and the \ac{MIMIC} framework that motivate the present work.

A particularly relevant body of work for this thesis already exists within the Silicon Synapse Lab. The foundational platform, \ac{B2}, was introduced by Ramezani et al.~\cite{ramezani2017batbot} as a 93~g autonomous robot with a carbon fiber skeleton, silicone membrane wings, and five degrees of actuation mimicking the dominant \ac{DOF}s of bat flight: shoulder retraction-protraction, elbow flexion-extension, wrist abduction-adduction, and dorsoventral leg movements. This work, published as the cover article in \textit{Science Robotics}, demonstrated the first autonomous flight of a bat-morphology robot performing banking turns, swooping, and straight flight through asymmetric wing folding. The preliminary untethered flights and design rationale for \ac{B2} were first reported in Ramezani et al.~\cite{ramezani2016icra} at \ac{ICRA} 2016.

A key challenge in replicating bat flight is the high dimensionality of wing kinematics. Hoff et al.~\cite{hoff2016synergistic} addressed this through a synergistic design approach using \ac{PCA} to extract dominant principal components from biological bat motion capture data, then optimizing \ac{B2}'s actuator trajectories through constrained optimization to match bat kinematic synergies. This work, presented at \ac{RSS}, established a principled methodology for reducing the actuation complexity of bat-inspired robots while preserving biologically meaningful morphing specializations. Hoff et al.~\cite{hoff2018optimizing} further optimized the structure and movement so that wing folding occurs at the correct phase in the wingbeat cycle, reducing actuation from five to three \ac{DOA}s and producing an average 89\% net lift improvement over the non-folding configuration, as published in the \textit{International Journal of Robotics Research}. The concept of describing bat robot flight through stable periodic orbits was explored by Ramezani et al.~\cite{ramezani2017orbits}, who demonstrated that locating feasible limit cycles and designing controllers to retain trajectories within bounded neighborhoods is an effective stabilization approach. Hoff et al.~\cite{hoff2019trajectory} subsequently developed trajectory planning methods using direct collocation for dynamically feasible bat-like flight maneuvers.

The subsequent platform, Aerobat, represents a significant evolution. Sihite et al.~\cite{sihite2020armwing} designed a bio-inspired articulated armwing structure using monolithic PolyJet \ac{3D} printing, combining rigid and flexible materials to form compliant linkage mechanisms that capture both plunging and flexion-extension flapping modes. Sihite and Ramezani~\cite{sihite2020cdc} derived a multi-body dynamical model of Aerobat capturing biologically meaningful \ac{DOF}s and demonstrated aerial body reorientation for perching maneuvers by enforcing nonholonomic constraints at \ac{IEEE} \ac{CDC} 2020.

The \ac{MIMIC} framework, proposed by Sihite et al.~\cite{sihite2021mimic}, integrates small low-power ``primer'' actuators with computational structures to achieve flight control through morphology changes rather than relying solely on closed-loop feedback. Under this paradigm, the computational structure encodes the desired wing kinematics in its geometry, and primer actuators introduce small, targeted modifications to that geometry to modulate the aerodynamic output. The thrust regulator mechanism developed in this thesis is a direct instantiation of the \ac{MIMIC} philosophy: a piezoelectric slip-stick actuator serves as the primer actuator, and the variable-length first radius link serves as the geometric modification point within the computational structure. This connection between the \ac{MIMIC} concept and the specific mechanism design presented in Chapter~\ref{chap:Methodology} represents the primary intellectual contribution linking the present work to the broader research program of the lab. Ramezani and Sihite~\cite{ramezani2022aerobat} further developed the actuation framework for Aerobat's 14 body joints, demonstrating the feasibility of joint motion regulation in untethered flights. Sihite et al.~\cite{sihite2022bangbang} achieved the first closed-loop stable flights of Aerobat using bang-bang control with horseshoe vortex shedding and Wagner function aerodynamic models.

The flagship journal publication by Sihite and Ramezani~\cite{sihite2024ijrr} in the \textit{International Journal of Robotics Research} introduces a morphology-centric \ac{MAV} design framework, demonstrating stable untethered dynamic morphing flights and aerodynamic force enhancement through wing surface maximization and rise-up time minimization. To address the hovering instability inherent in tailless morphing wing flight, Dhole et al.~\cite{dhole2023guard} developed a guard system with manifold small thrusters to stabilize Aerobat's position and orientation, creating a flapping-multi-rotor tandem platform.

Recent work includes banking turn controllers using collocation-based optimization~\cite{gupta2024banking}, conjugate momentum-based external force estimation~\cite{pitroda2025conjugate}, neural network and physics-based aerodynamic force estimation during tethered flight~\cite{gupta2025forces}, and \ac{ATMO}, a multi-modal morphing robot with ground-aerial transition capability~\cite{mandralis2025atmo}.

\section{Identified Research Gap}
\label{sec:gap}
The Silicon Synapse Lab's work on Aerobat demonstrates that multi-joint morphing wing flight is achievable with small, low-power actuators integrated into computational structures. Based on the solutions for lightweight actuators presented in the current research, this thesis focuses on the validation of several potential actuation methods for Aerobat. The gap in existing research that this thesis addresses is the integration of an actuation method within the mass and size constraints of the Aerobat platform. The following chapter describes the methodology used to validate the linkage-length modulation concept experimentally and to design, fabricate, and evaluate the actuated regulator mechanism.

\nocite{*}

% Methodology
\chapter{Methodology}
\label{chap:Methodology}

This chapter describes the methodology used to validate the concept of linkage-length-based thrust regulation and to design a mechanism capable of achieving this regulation on the Aerobat platform. Section~\ref{sec:aerobat_platform} introduces the Aerobat Delta test platform, its computational wing structure, and the control limitation that motivates this work. Section~\ref{sec:wing_regulation} describes how modifying the effective length of the first radius link $R_1$ alters the flapping gait and thereby modulates thrust. Section~\ref{sec:static_testing} presents the static regulator experiments using fixed-length \ac{FDM}-printed radius links. Section~\ref{sec:performance} derives the performance requirements for the regulator mechanism. Sections~\ref{sec:actuator_integration} and~\ref{sec:piezo_design} describe the iterative actuator selection process, progressing from string-tension and micro-servo approaches to the final piezoelectric slip-stick regulator design.

\section{Aerobat Flapping Wing Aerial Platform}
\label{sec:aerobat_platform}

\begin{figure}
    \centering
    \includegraphics[width=1\linewidth]{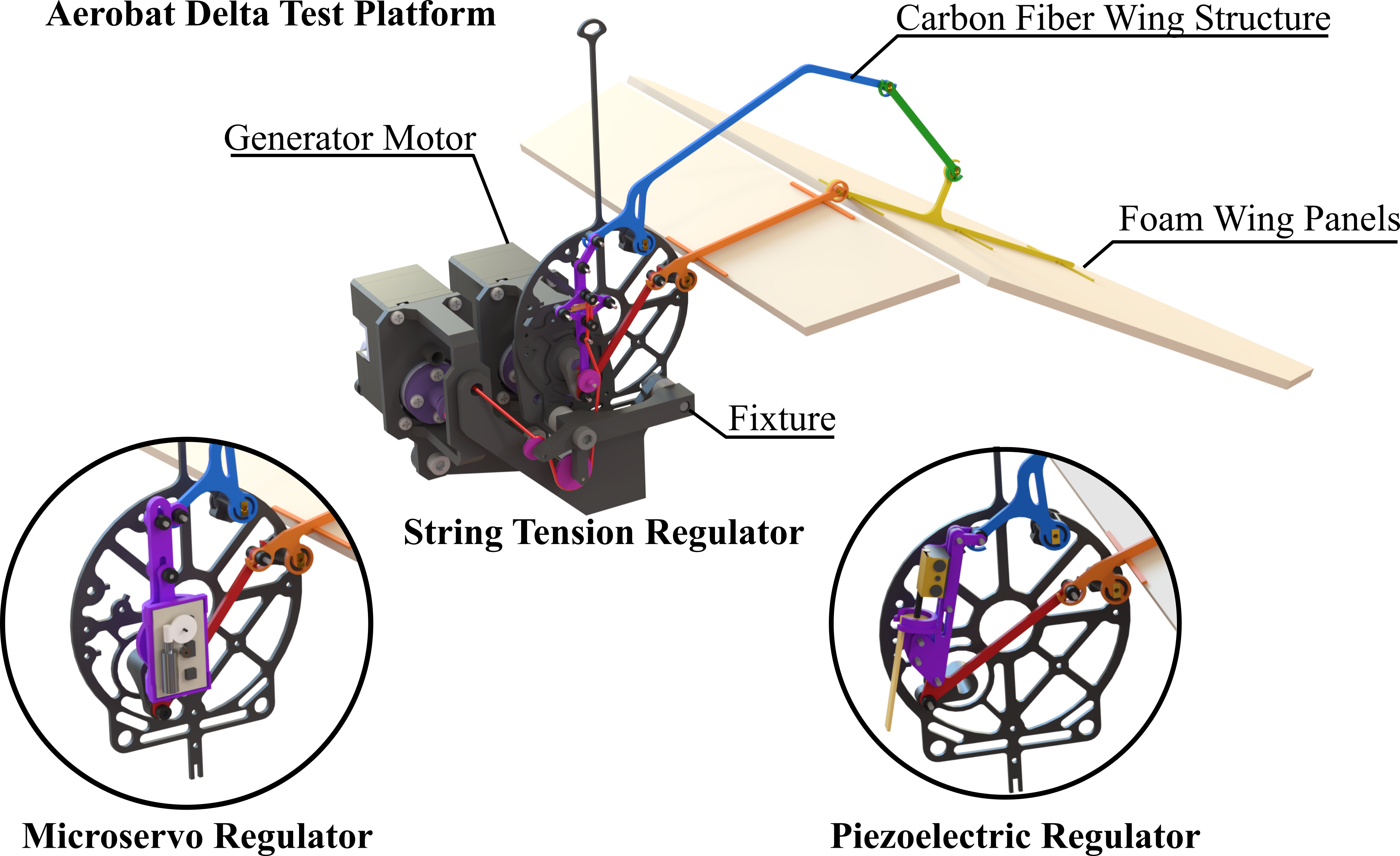}
    \caption{Aerobat Delta test platform with three regulator actuation methods}
    \label{fig:aerobat_delta_full_1}
\end{figure}

The primary structure of Aerobat is composed of several rigid links connected by revolute joints. Every link in the Aerobat version used in this study is fabricated from 1~mm-thick sheets of carbon fiber composite which is cut to shape via a \ac{CNC} router. These links are assembled in a planar manner, with stacked layers of the carbon fiber composite pieces joined by steel pins and \ac{FDM}-printed plastic retaining rings. In addition, several key joints where links are anchored to the main body of the aerial platform are reinforced with additional \ac{FDM}-printed plastic structures and brass rivets.

\begin{figure}
    \centering
    \includegraphics[width=0.75\linewidth]{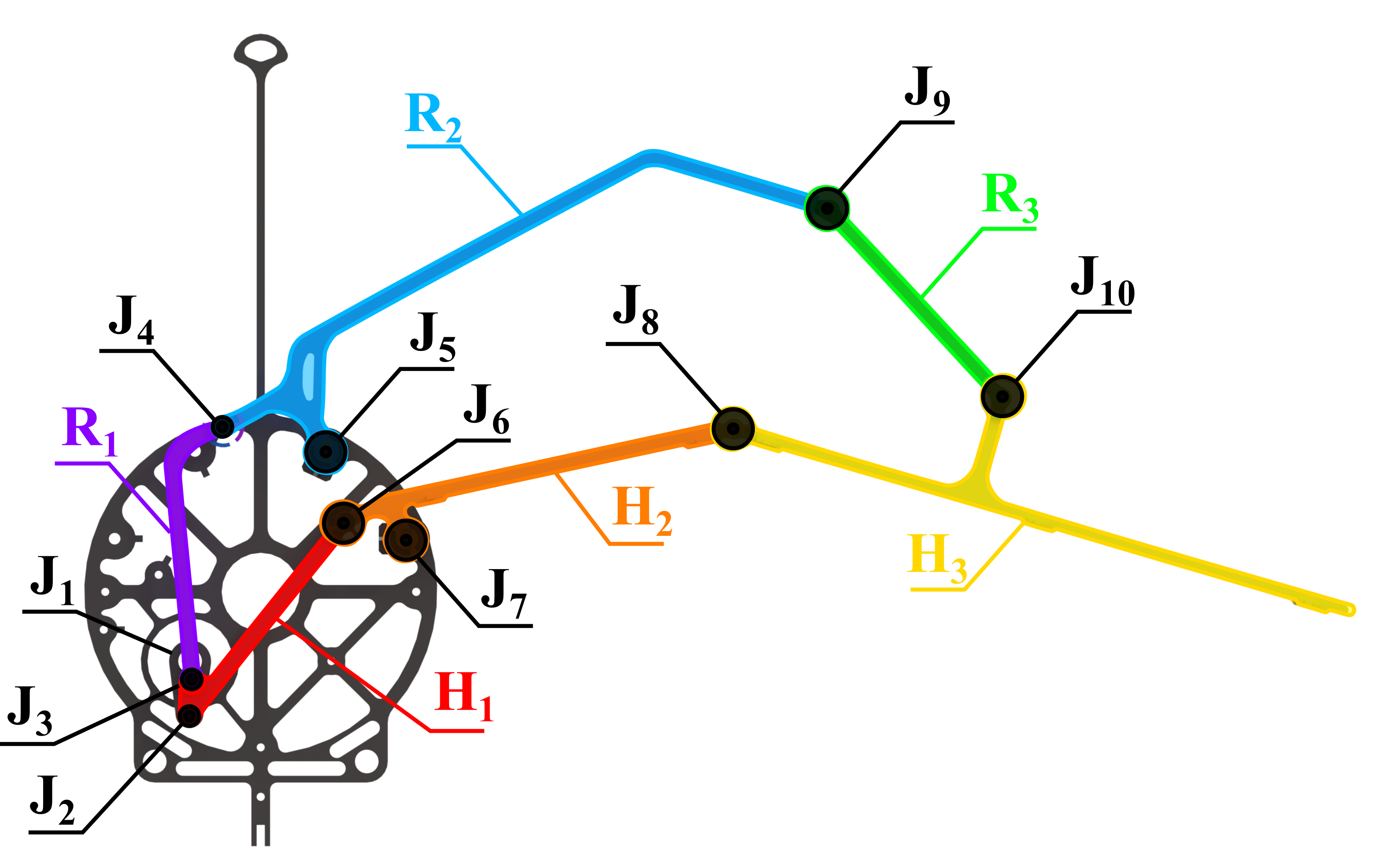}
    \caption{Aerobat Delta computational kinematic Structure linkages ($J_1-J_{10}$ are joints, $R_1-R_3$ are radius links, $H_1-H_3$ are humerus links)}
    \label{fig:aerobat_delta_front_1}
\end{figure}

The arrangement of these links and joints constitutes what is referred to as a computational structure~\cite{sihite2021mimic, sihite2024ijrr}. The link lengths, joint positions, and kinematic constraints of the mechanism are computationally optimized such that a single rotary input at joint $J_1$, driven by a \ac{BLDC} motor through a gear train, produces a complex, biologically meaningful flapping gait consisting of wing plunging, elbow flexion-extension, and membrane morphing. The geometry of the structure encodes the desired wing kinematics, eliminating the need for dedicated actuators at each joint. Figure~\ref{fig:aerobat_delta_front_1} shows the full linkage topology, with joints $J_1$ through $J_{10}$, radius links $R_1$ through $R_3$, and humerus links $H_1$ through $H_3$.

The primary issue this study addresses is the integration of a lightweight actuation method into Aerobat for the purpose of modulating the thrust produced on a per-wing level. While the computational linkage structure allows for dynamic wing motion to be generated by a single \ac{BLDC} motor, it does not allow for any thrust regulation or vectoring beyond altering the speed of both wings simultaneously. The motor speed is the only control input available, and it affects both wings symmetrically. The identified solution to this issue was the integration of additional, independently driven joints in the computational structure. For the initial prototype and proof of concept, a single mechanism was intended to be integrated into the first radius link $R_1$ of each wing. Altering the effective length of this link would affect the overall thrust output of a single wing, allowing for roll authority during flight.

While the goal of this study was to design a mechanism to regulate the thrust produced by each wing on Aerobat, it had several other critical concerns. The first, and most important, of these was weight minimization. The current Aerobat prototype has a measured mass of approximately 25~g, while the wings produce approximately 35~g of thrust, leaving a payload margin of roughly 10~g for all additional onboard systems. Any mechanism added to the platform must fit within this narrow margin. This weight constraint imposed restrictions on the regulator mechanism design, as it had to be constructed out of lightweight materials and contain as few components as possible. It also restricted the forms of actuation that could be used to those with a sufficiently high power-to-weight ratio for the given scale.

\section{Wing Computational Structure Regulation}
\label{sec:wing_regulation}

The proposed solution to the lack of control authority and stability in flight is the implementation of a thrust-altering mechanism in the wing structure. The existing computational structure is composed of rigid links and revolute joints that generate a consistent flapping gait when actuated via a single motor. As both wing halves are linked with a gear system, no independent control is possible. The mechanism outlined in this research is embedded in the first radius link $R_1$ on each wing of the aerial platform. This mechanism replaces the original rigid link with two shorter links connected by a prismatic joint, and is actuated to control the effective length.

The first radius link $R_1$ was selected as the target for length regulation because of its position in the kinematic chain. As shown in Figure~\ref{fig:aerobat_delta_front_1}, $R_1$ connects the motor-driven shaft at joint $J_1$ to the junction at $J_4$, where it interfaces with the remainder of the wing linkage. Because $R_1$ is the first link in the chain after the rotary input, changes to its effective length propagate through all downstream joints and links, producing a proportionally large effect on the overall wing trajectory. This makes $R_1$ the most kinematically influential point at which to introduce a geometric modification.

\section{Static Regulator Testing}
\label{sec:static_testing}

The preliminary step in designing this mechanism was to confirm the effects of changing the length of the selected wing structure link on the flap gait and lift produced. In order to test this, an experiment was set up with an existing Aerobat prototype. The target regulator actuation distance was then selected based on the results of this experiment. The lift force difference measured between the various regulator lengths produced a target displacement of 1.5~mm. The results of the experiment and the lift force measured are discussed further in Section~\ref{sec:static_lift_results}.

\section{Performance Characteristics}
\label{sec:performance}

In order to design a suitable length-changing mechanism, the expected load and desired change in length were required. Based on the \ac{FDM}-printed radii that were used in the experiment, a length change of at least 1.5~mm was determined to sufficiently alter the gait and lift force. The expected force load on the radius link required further work to derive. To ensure proper function of the regulator mechanism during both regular flight and aerial maneuvers, it was assumed that the wings would generate approximately 75\% of the maximum thrust during level flight. For maneuvers, a thrust generation of 100\% would be used. This additional thrust production is represented by $T_{max}$ in Eq.~\ref{eq:lift_force_1}, with a value of 1.33. The lift force produced by a single wing was isolated by halving this. A factor of safety of 2 was included to ensure proper functioning for all unforeseen loading circumstances. The target flight mass of Aerobat, including the bare platform, regulator mechanism, and payload, was estimated at 0.035~kg, and the gravitational acceleration was assumed to be 9.81~m/s$^2$. These values were used in Eq.~\ref{eq:lift_force_1} to calculate the predicted single-wing lift force, resulting in a value of 0.4566~Newtons.

\begin{equation}
    \label{eq:lift_force_1}
    F_L  = M_{Aerobat} * g * T_{max} * \tfrac{1}{2} * FoS
\end{equation}

\begin{figure}
    \centering
    \includegraphics[width=0.75\linewidth]{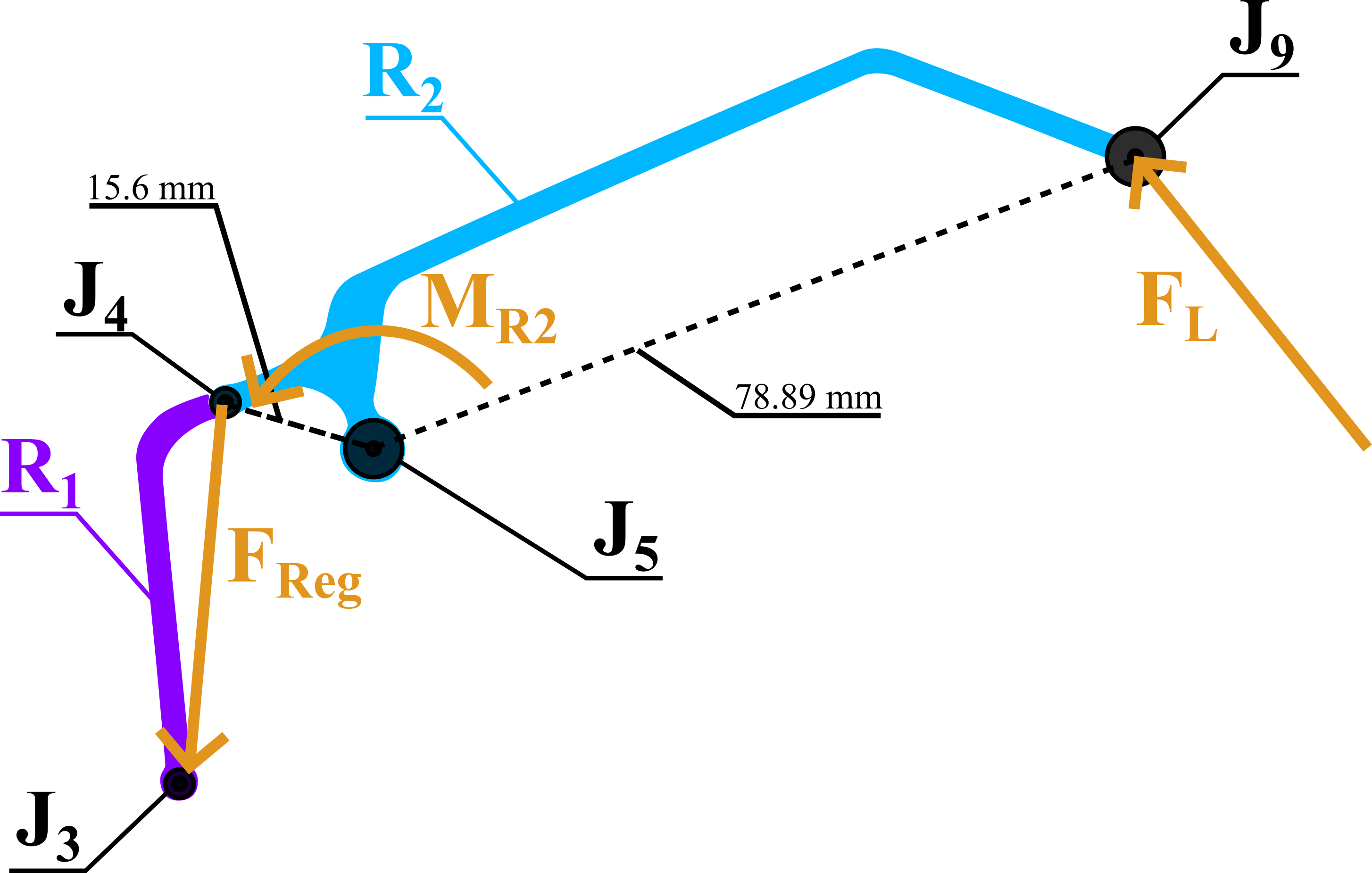}
    \caption{Aerobat Delta radius links $R_1$ and $R_2$ Free Body Diagram, with wing lift force $F_L$, $R_2$ moment $M_{R2}$ about $J_5$, regulator loading force $F_{Reg}$, and moment arm lengths}
    \label{fig:aerobat_delta_front_2}
\end{figure}

To isolate the force loading the regulator mechanism in the first radius link $R_1$, it was assumed that the entire lift force generated by the wing was acting upon it. This was assumed as the worst-case loading scenario, and is likely much lower during normal operation. Based on the geometry of the connected radius links and base plate, a lever and fulcrum mechanical advantage was calculated. This is the quotient of the link lengths on either side of the fulcrum joint. As shown in Fig.~\ref{fig:aerobat_delta_front_2}, the second radius link $R_2$ creates a lever about shoulder joint $J_5$ with moment arm lengths of 78.89~mm and 15.6~mm on each respective side. These distances were used in Eq.~\ref{eq:reg_force_1} and applied to the previously derived single-wing peak lift force $F_L$, resulting in a predicted force of 2.309~N, or 235.45~grams-force, loading the regulator mechanism along the axis of motion.

\begin{equation}
    \label{eq:reg_force_1}
    F_{Regulator} = \frac{d_{J_5,J_9}}{d_{J_5,J_4}} * F_L
\end{equation}

The worst-case assumptions in this calculation were made due to the difficulty in accurately analytically deriving the actual portion of the lift force experienced as load by the $R_2$ radius link. In addition to not having a way to solve for this directly, the actual lift force generated by each wing during level flight was unknown. The required performance parameters that resulted from the static regulator testing and peak loading derivation are outlined in Table~\ref{table:performance_requirements}.

\begin{table}
\centering
\caption{Derived performance requirements for a viable regulator mechanism design}
\label{table:performance_requirements}
\begin{tabular}{ |c|c|  }
 \hline
    Parameter & Value\\
 \hline
 Displacement & 1.5~mm\\
 \hline
 Actuation Force & 235.45~gf\\
 \hline
\end{tabular}
\end{table}

In addition to the performance parameters outlined previously, several other constraints were imposed upon the mechanism design. The first of these was the limitation of material choices and fabrication methods available to the Silicon Synapse Lab. The primary method of fabrication used throughout this thesis was \ac{3D} printing. This was done with both \ac{FDM} and \ac{SLA} printers with a range of materials. These were utilized due to the relatively low cost per unit material and ease of access for members of the Silicon Synapse Lab. As the intent behind this thesis is the fabrication and testing of a flight-ready regulator mechanism to interface with the existing structure of Aerobat, all versions used in this study were designed with the intent of eventually being fabrication from \ac{CNC}-cut carbon fiber composite sheets. However, for the purposes of testing, \ac{3D} printing was the primary method of fabrication. To minimize weight and component complexity, all mechanism prototypes were also designed to utilize existing steel pins as the fastening method. This consisted of small 5-10~mm segments of 1~mm diameter steel wire and a friction fit between the pins and plastic components. This was selected due to the availability of existing steel pins in the Silicon Synapse Lab inventory and ease of manufacturing of new pins if required. Actuators selected for study in this thesis were also chosen with the availability as a primary driving factor. All motors used were done so due to their availability to the Silicon Synapse Lab.

With the displacement and force requirements established, along with the material and fabrication constraints, the next phase of the project focused on selecting an actuation method and designing a variable-length mechanism that could meet these targets while remaining within the approximately 10~g payload margin of the platform. The following sections describe the three actuation approaches that were evaluated: string-tension, micro-servo, and piezoelectric slip-stick.

\section{Regulator Actuator Integration}
\label{sec:actuator_integration}

With the design constraints and performance targets set, the second phase of this project was to design a variable-length mechanism to be embedded in the radius link. Although the \ac{FDM}-printed radius link replacements provided a set of fixed lengths for the initial experiment, they did not allow for changing the length during motion. Direct control over the length of the regulator and the ability to change it during flight was required to fulfill the intent of this study. To this end, several actuation methods were considered. These were selected on the basis of the materials and manufacturing methods available to the Silicon Synapse Lab, in addition to ease of control and simulation. Prototypes of each actuation method were manufactured and tested to determine the viability of each before settling on a final system. 

\subsection{String Tension Actuation}

\begin{figure}
    \centering
    \includegraphics[width=0.75\linewidth]{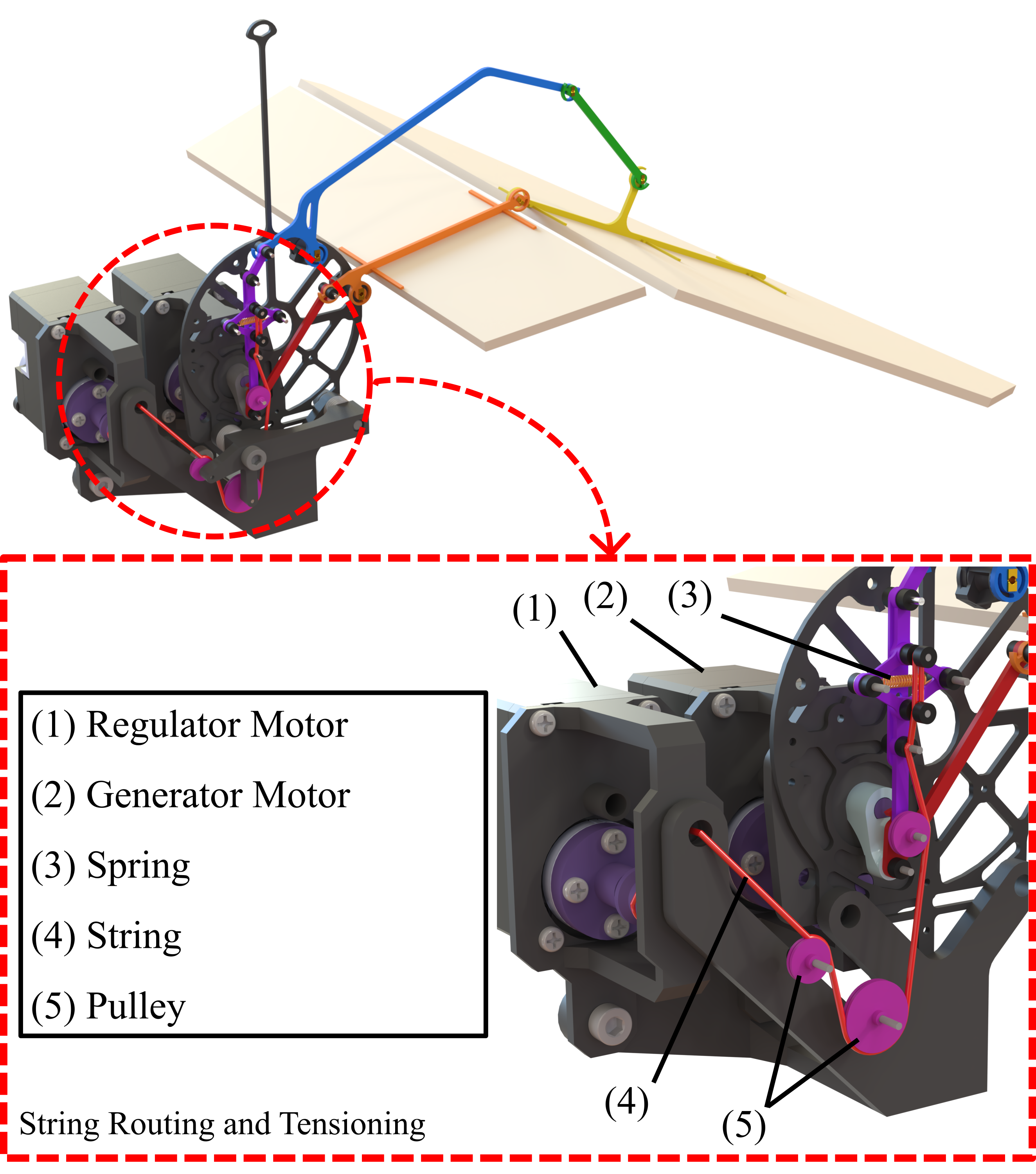}
    \caption{String-tension actuated regulator assembly. top: full placement in Aerobat Delta, bottom: string routing path and components}
    \label{fig:string_regulator_exploded}
\end{figure}

After confirming the effect of the $R_1$ link length on the lift force generation, the next step was to test an actuated variable-length mechanism. The initial regulator design utilized a string and a set of pulleys to drive a spring-loaded mechanism. The string was routed around the set of pulleys and into a motor-driven spool. The pulleys were attached to several key joint axles in the existing Aerobat Delta computational structure. In addition to the embedded pulleys, several others were attached to the \ac{3D}-printed mounting bracket to proper string routing and low friction. The length of the regulator mechanism was controlled by balancing the tension of the string and spring. The components and layout of the spring-loaded regulator mechanism is shown in Fig.~\ref{fig:string_regulator_exploded}. The motors selected to drive this mechanism were the same Dynamixel servomotors as the static regulator experiment. Similar to the previous tests, these were again selected for the output torque and rotational speed potential. Additionally, the sensors integrated into each motor were critical for not only the data collection but the actuation of the regulator mechanism. Synchronization of the string tension and driveshaft angle was required due to the changing length of the string route throughout the course of each flap. The two motors were synchronized via the collected angle and velocity data from each servo. This was handled by a Python script controlling both motors simultaneously. The angular position of the regulator motor was calculated as a function of the measured angle of the generator motor. Prior to testing, the system was calibrated and angles were measured 16 times across a flap cycle. This provided a calibration curve that could be followed by the regulator motor based on the position of the generator motor.

While this setup was able to function while the wing was stationary, it suffered from several issues preventing proper static or dynamic testing. The first issue discovered was the tolerance of the pulleys and axles. These were loose in the structure, and allowed the axis of motion to tilt relative to the rest of the robot. This effect was further enhanced by the inherent flexibility of the carbon fiber composite linkages. The deformation of this structure allowed the string tension affecting the regular mechanism to deviate significantly from the tension applied by the motor. This led to severe under-actuation as the deformation required less force than the proper actuation of the regulator did. These issues were compounded by the weakness of the string. The choice of nylon line for this application led to additional flexibility and tearing, especially while under tension for extended periods of time. 

In addition to the mechanical issues with this approach, there were also several hardware limitations. The maximum output speed of the tension motor was not high enough to synchronize with the main motor. This was due to the use of a gearbox on the main motor, which provided a 2:1 speed increase. The lack of a second gearbox prevented the tension motor from fully matching the targeted speed and position. Further issues were discovered while using the Dynamixel servomotors. While the regulator motor followed a calibrated trajectory as a function of the generator motor position, it tended to deviate significantly between tests. This required recalibration between each flapping test. Any minor inaccuracy in this calibration led to further issues with flexing and string breakage during each subsequent test as components were pulled out of position. 

The string-tension design, despite its failure during dynamic testing, provided two important design insights that informed subsequent iterations. First, any regulator mechanism that relies on force transmission through the existing computational structure, whether via strings, pulleys, or other external routing, is inherently vulnerable to the compliance and tolerance stack-up of the carbon fiber linkage. This pointed toward a self-contained mechanism embedded directly within the target link, rather than one actuated remotely through the structure. Second, the synchronization requirement between the regulator motor and the flapping motor introduced significant software and hardware complexity that would be difficult to manage reliably in flight. A simpler actuation approach with fewer inter-dependencies was needed. These observations motivated the evaluation of sub-gram micro-servo motors, which offered the potential for a compact, self-contained mechanism with direct position control.

\subsection{Micro-Servo Actuation}
Although the static wing performance of the string-driven regulator mechanism showed promise, it failed to function during flapping tests due to the reasons listed above. This failure led to the selection of a new actuation method: the sub-gram micro-servo motor. The low mass and \ac{PWM}-based position control made them an appealing choice for this application. According to the manufacturer, the line of sub-gram micro-servos selected was ideal for small \ac{RC} aircraft, which translated well to the Aerobat platform.

\begin{figure}
    \centering
    \includegraphics[width=12cm]{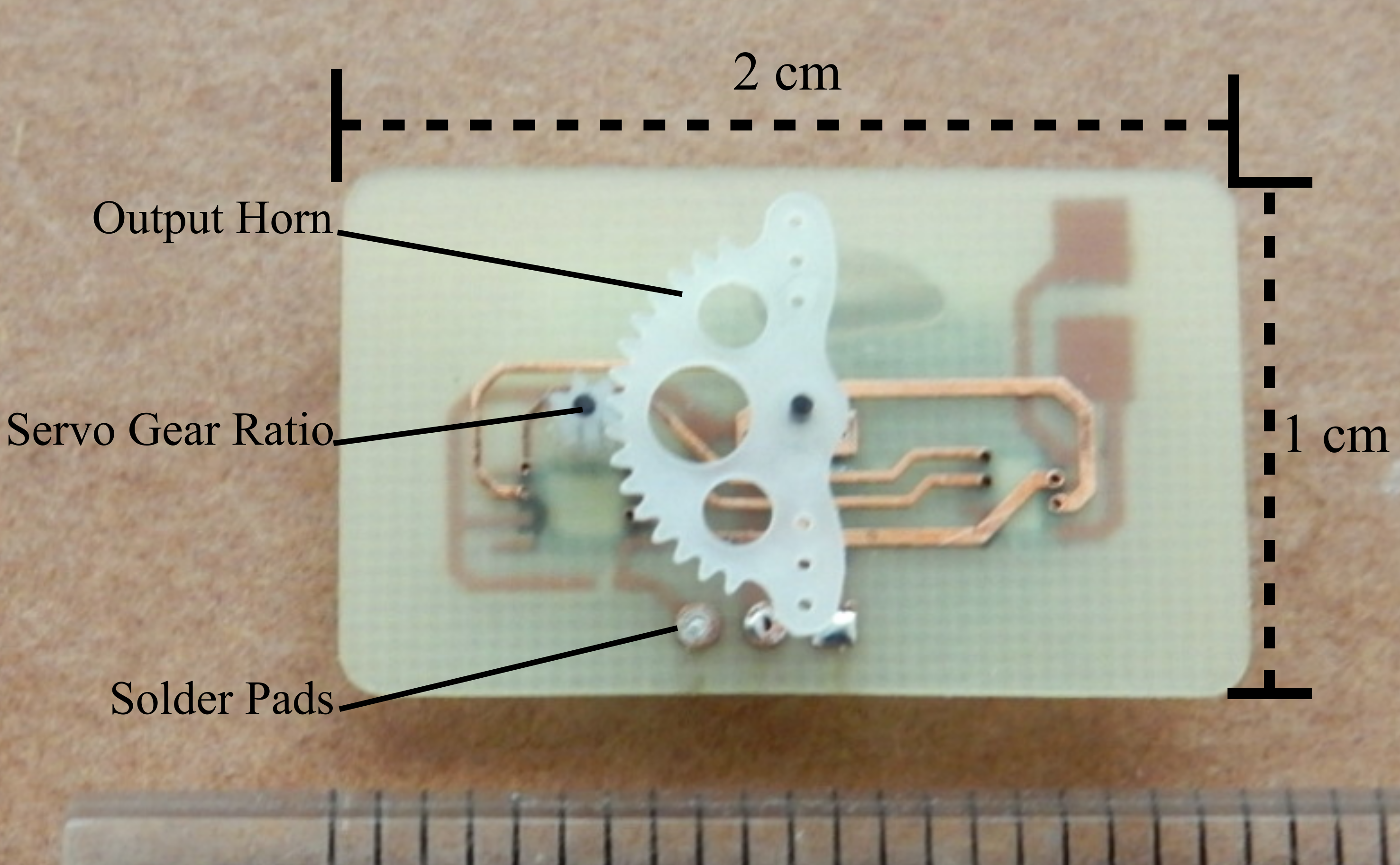}
    \caption{Sub Micro .50~g LZ Servo with components, adapted from \cite{Leichty}}
    \label{fig:microservo_product_image}
\end{figure}

\begin{figure}
    \centering
    \includegraphics[width=0.75\linewidth]{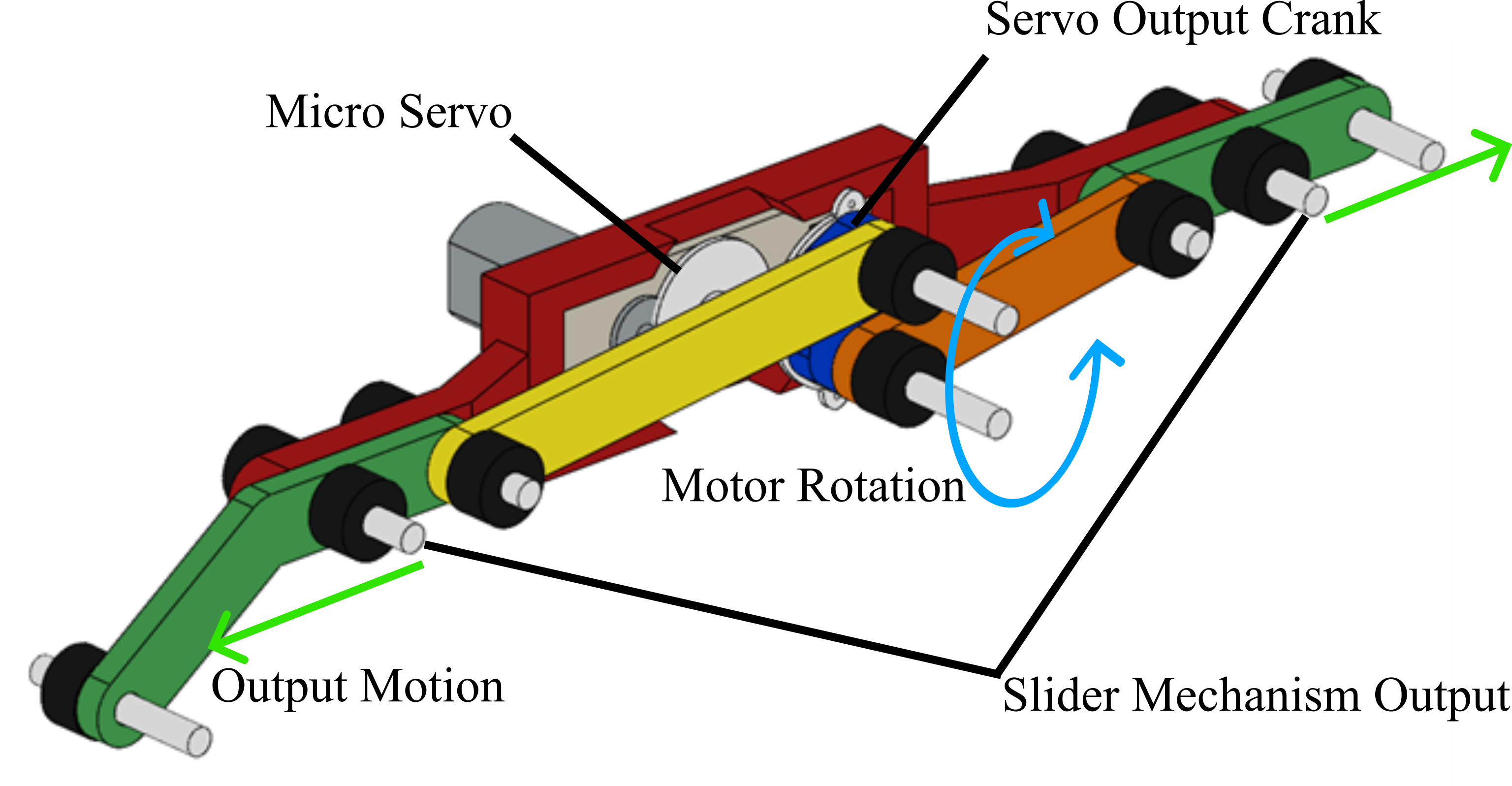}
    \caption{Micro-servo actuated regulator concept with motor action and motion output}
    \label{fig:microservo_idea_color}
\end{figure}

\begin{figure}
    \centering
    \includegraphics[width=1\linewidth]{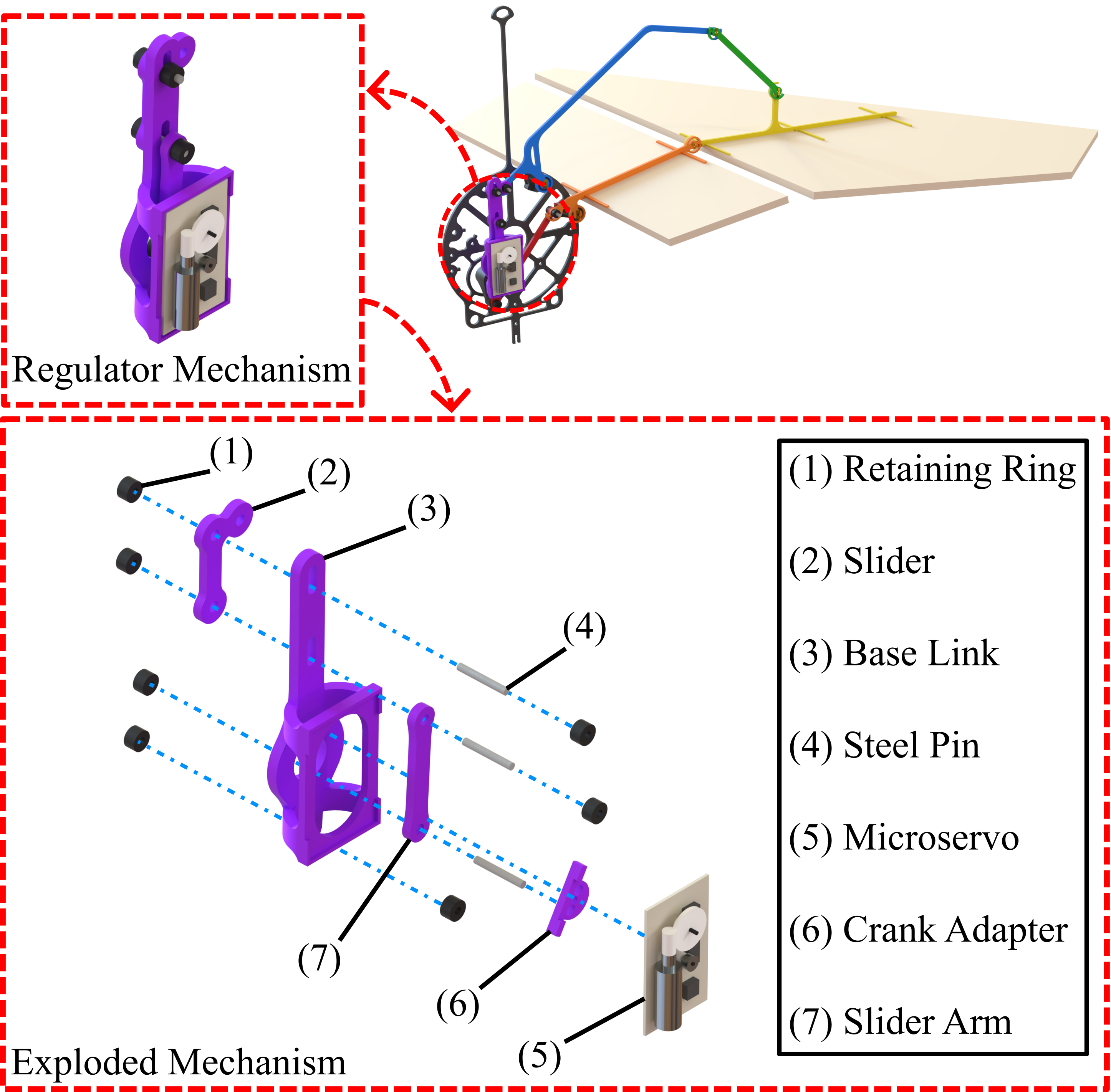}
    \caption{Sub Micro .50~g LZ Servo actuated regulator mechanism. top right: placement in Aerobat Delta, top left: regulator mechanism, bottom: regulator mechanism exploded view and components}
    \label{fig:microservo_CAD}
\end{figure}

The micro-servo model chosen for this thesis is the Sub Micro .50~g LZ Servo from Micro Flight Radio, shown in Fig.~\ref{fig:microservo_product_image}. This motor was to be integrated directly into the regulator mechanism due to the low mass of the motor and small footprint. The output shaft of the motor drove the linear motion of the regulator components via a slider-crank mechanism. The conceptual design for this mechanism is shown in Fig.~\ref{fig:microservo_idea_color}. As shown in Fig.~\ref{fig:microservo_CAD}, the final design of the mechanism slightly different, with a single larger slider-crank system replacing the dual collinear pair in the initial concept. This change was made due to difficulties with fastening the components together during fabrication. The output shaft adapter of the servo motor was too small at this scale to handle multiple 1.05~mm diameter joint pins, and prevented the crank and slider components from properly aligning. In addition this, the fragility of the motor \ac{PCB} and gears led to damaging several of the micro-servos, and required a redesign before further testing could take place.

Despite the changes made to the micro-servo-actuated regulator design, the final version also failed during testing. While increasing the size of all components prevented the links in the mechanism from breaking or the joint shafts from falling out as was seen during initial tests, the low structural integrity of the motors themselves was unsolvable. The board to which the circuitry, motor, and gear shafts were mounted were extremely delicate and prone to delamination during soldering. In addition, the small plastic gears tended to lose teeth when high torque loads were experienced. The circuitry on each board also appeared to be defective, not responding properly to \ac{PWM} signals as described by the documentation provided from the manufacturer. Due to these issues, the usage of sub-gram micro-servos was discontinued in this thesis. 

\section{Piezoelectric Regulator Design} 
\label{sec:piezo_design}

After the failure of the sub-gram micro-servos during preliminary tests, a final method of actuation was selected: slip-stick based piezoelectric motors. Specifically selected was the \ac{TULA} family of piezoelectric linear actuators, given the availability of a \ac{TULA}-50 motor in the Silicon Synapse Lab inventory. These utilize a piezoelectric crystal to vibrate the shaft at ultrasonic frequencies. By altering the duration of each pulse in the driving signal, the output fixture of the actuator can be made to slide linearly in either direction along the shaft. This family of actuators was chosen due to the extremely light weight, high power-to-weight ratio, and low energy consumption. This is a side effect of the slip-stick mechanism, as the output is held in place by static friction forces when not actuating. This produces a very high holding force relative to the overall strength of the actuator. 

\begin{figure}
    \centering
    \includegraphics[width=\linewidth]{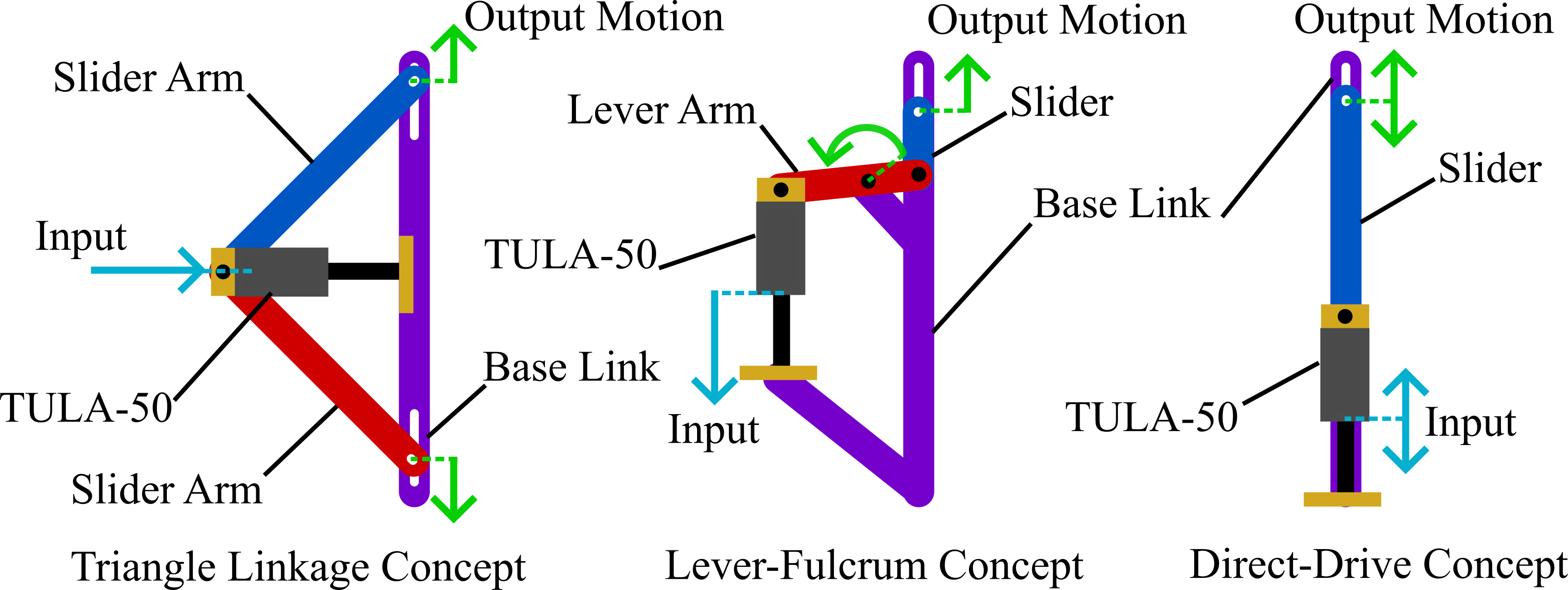}
    \caption{Piezoelectric regulator mechanism concepts. left: single triangle linkage, middle: lever-fulcrum linkage, right: direct-drive linkage, with basic component structure and mechanical movement}
    \label{fig:piezo_concept_sketches}
\end{figure}

Unlike the previous two actuation methods considered, the \ac{TULA} family of piezoelectric actuators are linear motors. While this greatly simplified the mechanism, the lower overall force of the actuators required a higher mechanical advantage in order to reach the target performance of 235.45~grams-force. In order to achieve this mechanical advantage, a more complex linkage system was required. Several initial proposals for this mechanism are shown in Fig.~\ref{fig:piezo_concept_sketches}. The concepts considered for this purpose include multiple variants of a triangle linkage, a lever-fulcrum system, and a direct drive. Each of these presented a unique set of design and implementation challenges and required careful consideration.

\begin{figure}
    \centering
    \includegraphics[width=1\linewidth]{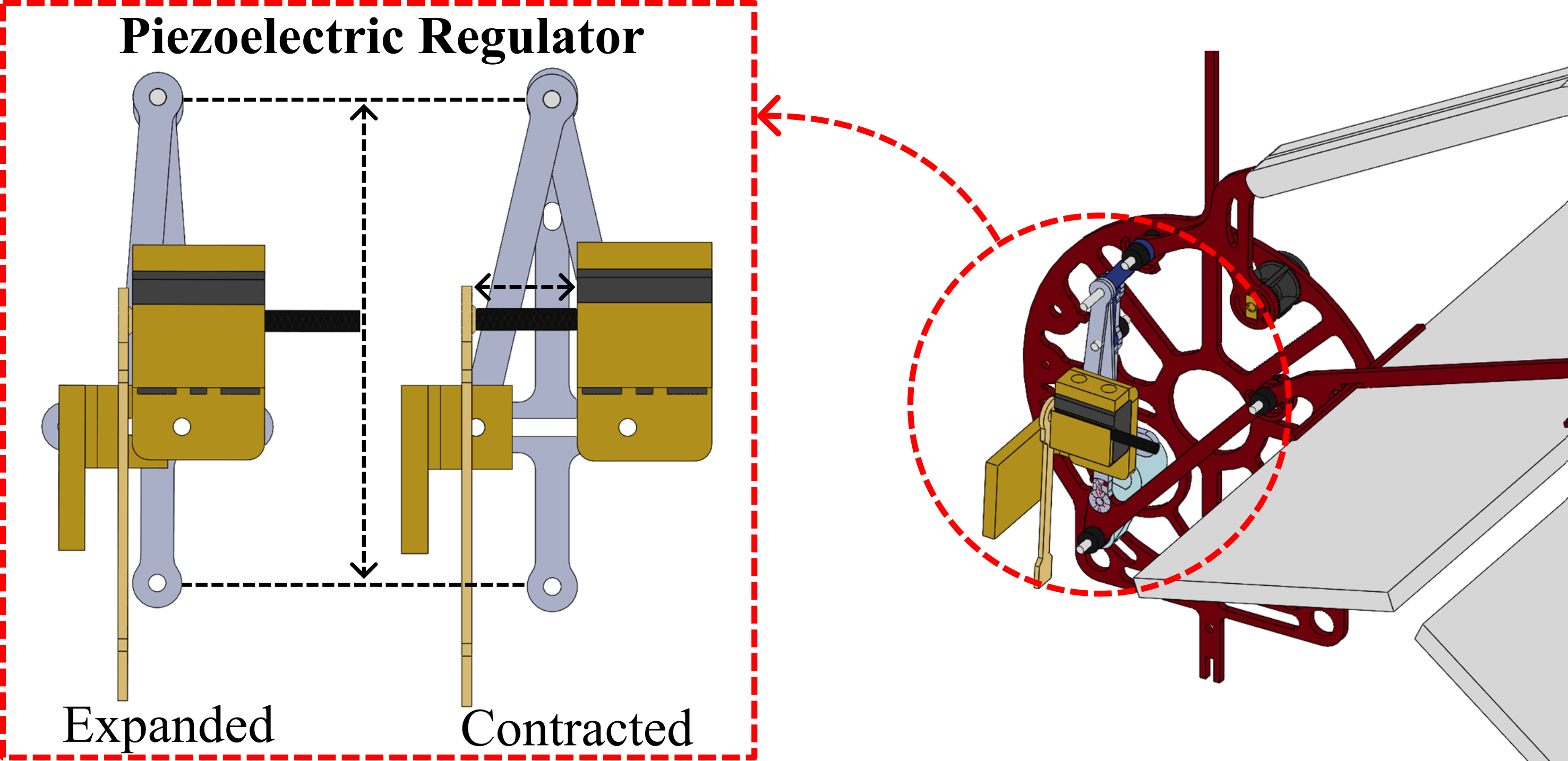}
    \caption{Initial piezoelectric single-triangle linkage regulator concept, with placement in Aerobat Delta}
    \label{fig:piezo_first_concept_CAD}
\end{figure}

The first concept considered was the single-triangle linkage. This concept had the highest potential mechanical advantage given the number of components required. While the lever-fulcrum system could achieve even higher degrees of mechanical advantage, the size of the components would have prevented fabrication. In light of this, the single-triangle mechanism was selected first. The initial concept for this mechanism is shown in Fig.~\ref{fig:piezo_first_concept_CAD}.

The first step in designing this system was to determine the link lengths required to produce the desired mechanical advantage. Given the force output of the \ac{TULA}-50 actuator while in motion, a minimum mechanical advantage of 4:1 was selected. As the piezoelectric motor had an actuation range of 6~mm, the mechanical advantage would have to remain at least 4:1 over the entire distance. This also guaranteed an output range of 1.5~mm, fulfilling the other regulator performance requirement. While determining the mechanical advantage necessary for this, the link lengths required became longer than the original concept design shown in Fig.~\ref{fig:piezo_concept_sketches} could accomodate. This resulted in the orientation change made to the conceptual design in Fig.\ref{fig:piezo_first_concept_CAD}. Based on the geometry in this mechanism design concept, a mathematical system was derived. In order to select the link lengths based on this ideal mechanical advantage, Eq.~\ref{eq:single_triangle_mechad_1} was used. This was derived from the right triangle geometry of the system.

\begin{equation}
    \label{eq:single_triangle_mechad_1}
    d_{output} = \frac{\tfrac{1}{2}(d_{input}+d_{inital}-L_{base})}{\sqrt{L_{hyp}^2 - \tfrac{1}{4}(d_{input}+d_{inital}-L_{base})^2}} d_{input}
\end{equation}

\begin{equation}
    \label{eq:single_triangle_mechad_2}
    IMA = \frac{d_{output}}{d_{input}}
\end{equation}

\begin{figure}
    \centering
    \includegraphics[width=12.5cm]{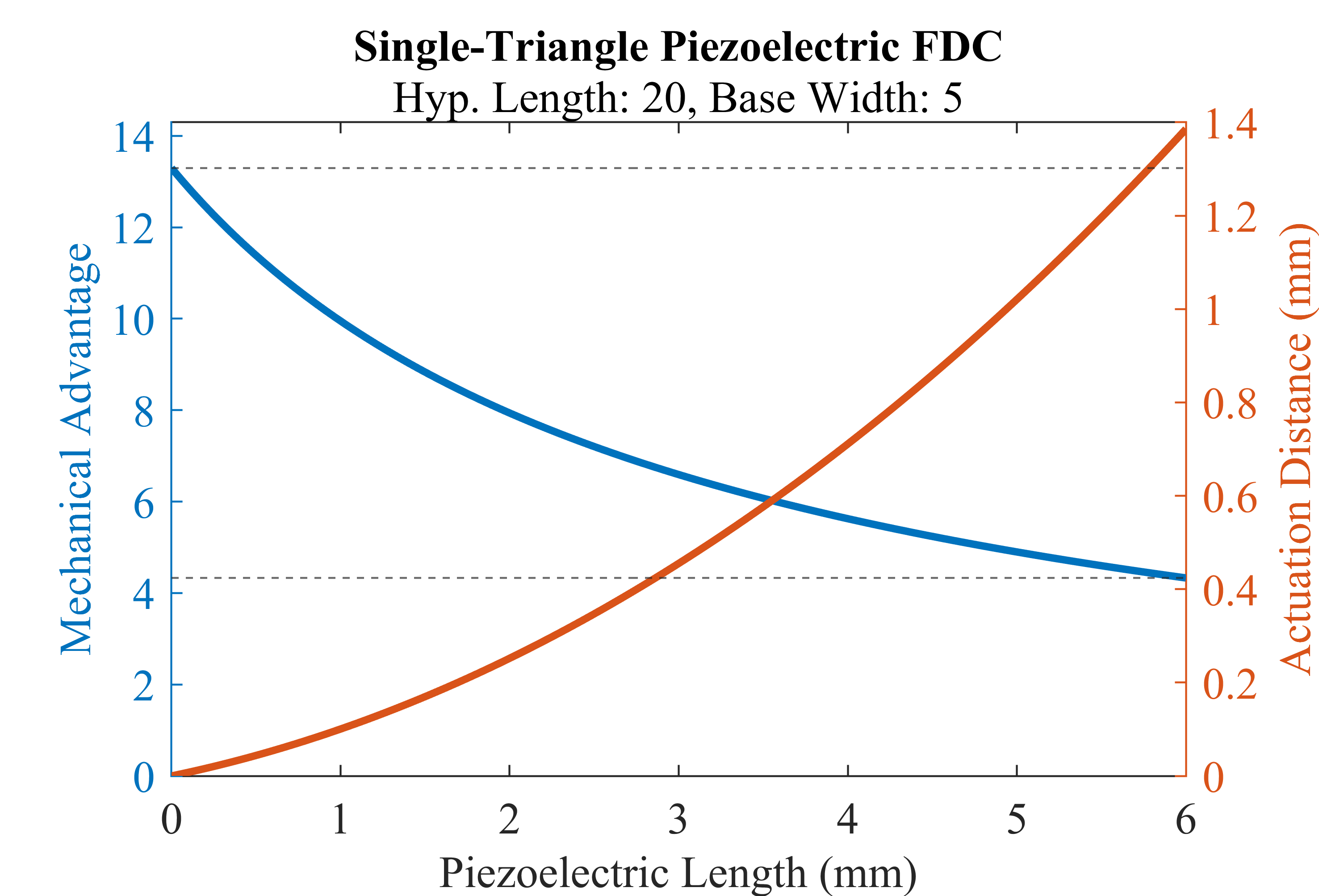}
    \caption{Predicted single-triangle piezoelectric performance across range of actuation, with Ideal Mechanical Advantage and output actuation distance}
    \label{fig:piezo_performance_plot}
\end{figure}

The output movement $d_{output}$ was calculated across the entire 6~mm actuation range of $d_{input}$. Using an initial length of 8~mm in $d_{initial}$, an offset $L_{base}$ of 5~mm, and a hypotenuse length $L_{hyp}$ of 20~mm, the maximum output movement distance was limited to under 1.4~mm. The ideal mechanical advantage of this system was then calculated using Eq.~\ref{eq:single_triangle_mechad_2}. The predicted mechanical advantage and output actuation distance over the range of motion for the \ac{TULA}-50 motor is shown in Fig.~\ref{fig:piezo_performance_plot}. As shown in the plot, the minimum mechanical advantage is greater than 4 and reaches over 13 near the start of the actuator stroke. While non-linear in nature, the input-output actuation distance relationship is sufficiently linear for predictable regulator functionality across the entire actuation range.

\begin{figure}
    \centering
    \includegraphics[width=12.5cm]{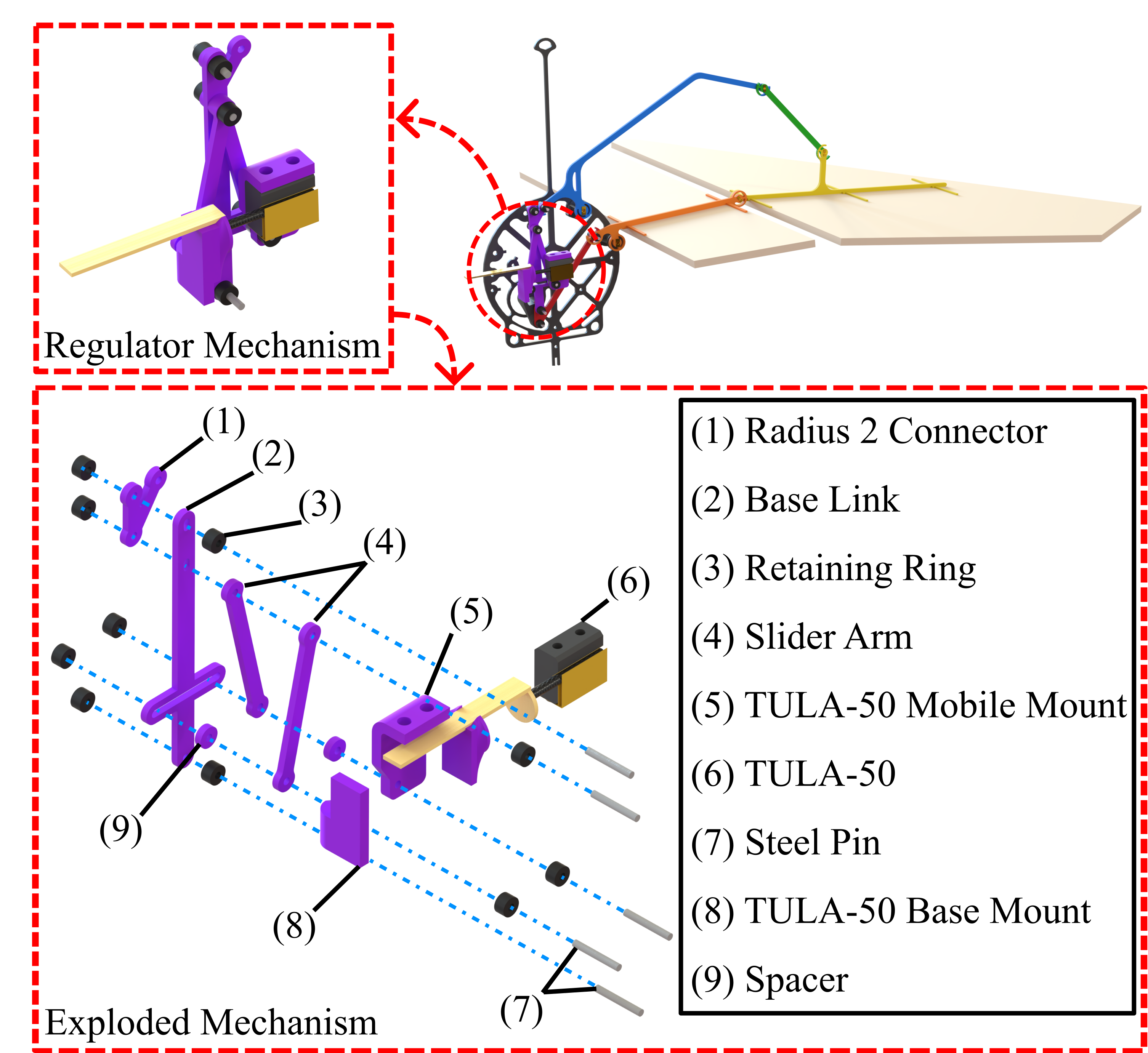}
    \caption{Single-triangle piezoelectric slip-stick actuated regulator mechanism. top right: placement in Aerobat Delta, top left: regulator mechanism, bottom: regulator mechanism exploded view and components}
    \label{fig:piezoelectric_triangle_regulator_CAD}
\end{figure}

Based on this analysis, a hypotenuse link length of 20~mm and base width of 10~mm was selected. This resulted in a regulator design that would perform according to the requirements set previously. This mechanism design is shown in Fig.~\ref{fig:piezoelectric_triangle_regulator_CAD}. The components for this prototype were fabricated using \ac{SLA}-printing techniques. This allowed for rapid prototyping while providing superior material properties when compared to \ac{FDM} printing. The most important among these material differences was the coefficient of friction, as the surface smoothness of \ac{SLA} resin components is far lower. In addition, the spatial resolution of the \ac{SLA} printers is significantly higher. This allowed for tighter part tolerances and less mechanical play. Despite these advantages, the prototype was unable to properly actuate.

Based on the performance of the initial single-triangle regulator design, a final set of changes was made. The direct-drive approach mentioned previously was implemented. This drastically reduced the number of components required for the mechanism. However, it removed the mechanical advantage provided by the previous design. While this direct-drive mechanism would not have the required mechanical advantage to actuate the wing during flight, it was selected in order to minimize the internal friction that prevented the previous prototype from moving. 

\begin{figure}
    \centering
    \includegraphics[width=12.5cm]{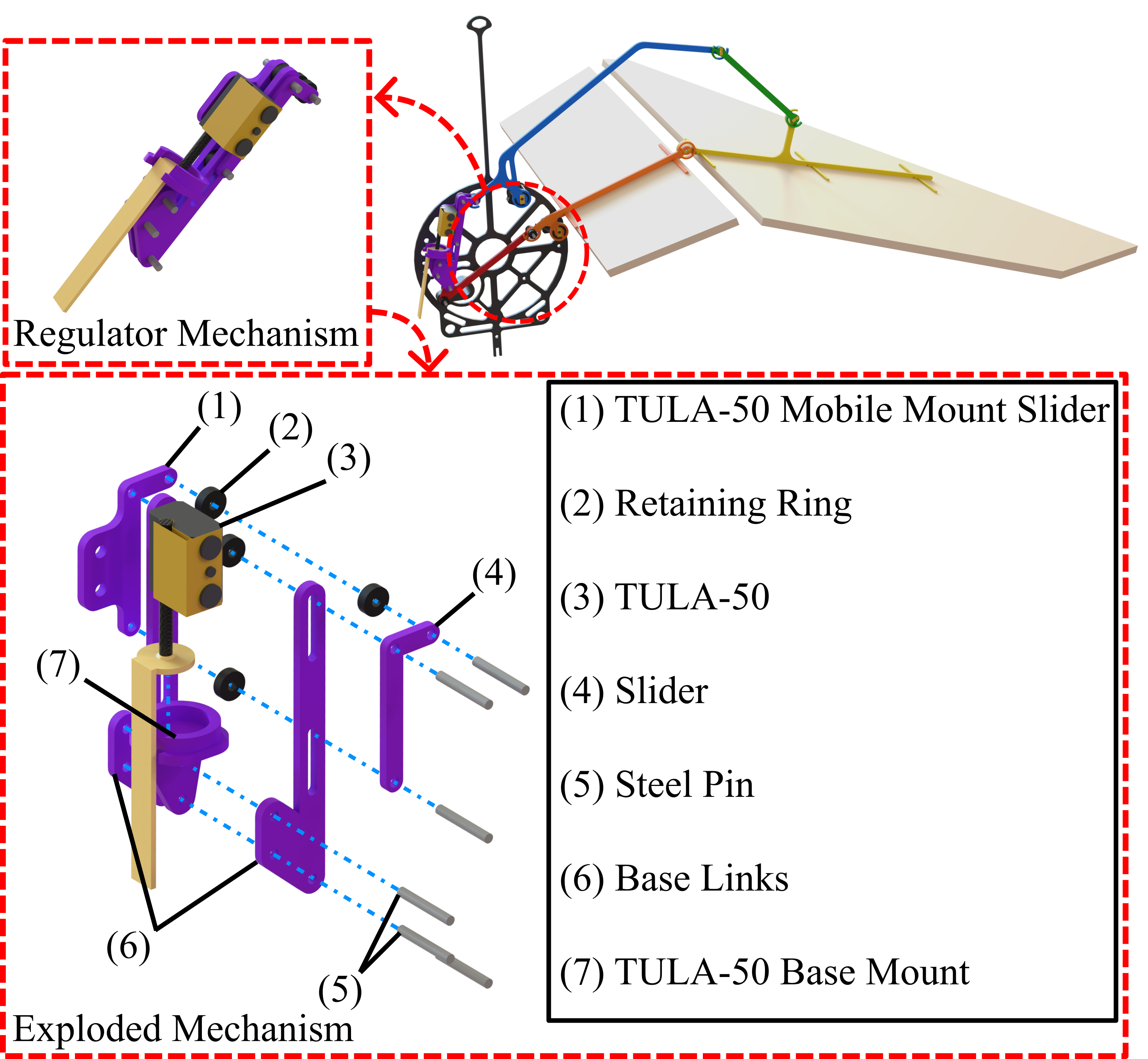}
    \caption{Direct-drive piezoelectric slip-stick actuated regulator mechanism. top right: placement in Aerobat Delta, top left: regulator mechanism, bottom: regulator mechanism exploded view and components}
    \label{fig:piezoelectric_regulator_CAD}
\end{figure}

This initial design for the direct-drive piezoelectric regulator was refined until a final version was reached. This was the result of multiple rounds of prototyping and rudimentary testing. The first fabrication attempt utilized \ac{FDM} printing and the previously mentioned steel pins. However, this attempt failed as the \ac{FDM}-printed components still produced a large amount of internal friction that prevented the actuator from functioning. This led to the decision to use \ac{SLA} printing instead, as that technology produced a lower surface coefficient of friction due to the smoother material and finer layer thickness. While this solved the internal friction problems encountered with \ac{FDM} printing, the resin components proved far too fragile once cured. This led to components breaking during assembly of the prototype. As a final attempt, \ac{FDM} printing was used in conjunction with manual component post-processing. By sanding the surfaces in contact with other components in the mechanism, the sliding friction of the system was greatly reduced. This allowed the \ac{TULA}-50 actuator to function properly and drive the regulator mechanism. The final design of the direct-drive piezoelectric regulator mechanism is shown in Fig.~\ref{fig:piezoelectric_regulator_CAD}.

% testing and discussion
\chapter{Results}
\label{chap:results}

\section{Static Lift Testing}
\label{sec:static_lift_results}
\begin{figure}
    \centering
    \includegraphics[width=12.5cm]{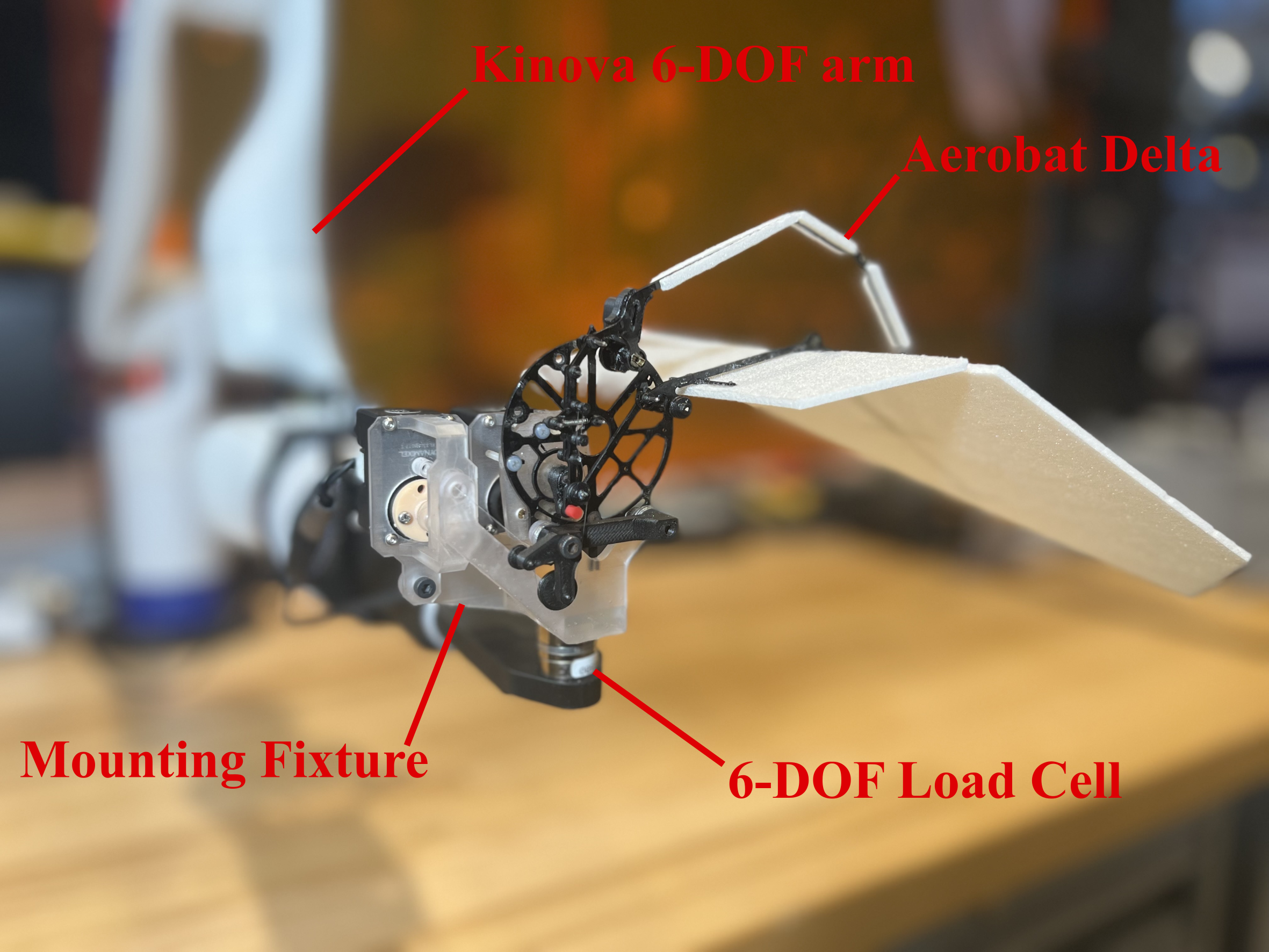}
    \caption{Experimental setup with the Aerobat Delta prototype on a 6-axis load cell and robot arm.}
    \label{fig:aerobat_on_arm_setup}
\end{figure}

The initial step in the design process of the regulator mechanism was to validate the effect of radius length on the flapping gait and lift production. This was done using the Aerobat Delta prototype, which consisted of the original Aerobat body with only a single wing. This was mounted to a fixture that anchored the aerial system to a 6-axis load cell on the end effector of a Kinova 6-\ac{DOF} Arm. The load cell was connected to a computer via a networking device that recorded the forces and torques on Aerobat Delta. The entire setup was fixed to a table in order to provide a stable environment without significant motion disturbances. Figure~\ref{fig:aerobat_on_arm_setup} shows the experimental setup of Aerobat Delta mounted to the load cell on the Kinova arm. 

The experimental procedures consisted of replacing the first radius link $R_1$ of Aerobat Delta with a set of \ac{3D}-printed copies at varying lengths. These were manufactured using \ac{FDM} printing, due to the ease of use and speed. The length of these radius copies was determined based on the length of the original $R_1$ link. Using the original length of 29.33~mm as a baseline, 3 copies were made. The lengths considered included the original 29.33~mm, and a difference of 0.75~mm both longer and shorter. This increment was selected to produce a total range of 1.5~mm across the three lengths, corresponding to the estimated stroke of a compact piezoelectric actuator that could be embedded in the link. The final lengths selected for testing are displayed in Table~\ref{table:icra_length_matrix}.

\begin{table}
\centering
\caption{Simulated Radius $R_1$ Lengths}
\label{table:traj_length_matrix}
\begin{tabular}{ |c|  }
 \hline
    Radius $R_1$ Length (mm) \\
 \hline
 28.58\\
 \hline
 28.77\\
 \hline
 28.96\\
 \hline
 29.14\\
 \hline
 29.33\\
 \hline
 29.52\\
 \hline
 29.71\\
 \hline
 29.89\\
 \hline
 30.08\\
 \hline
\end{tabular}
\end{table}

\begin{figure}
    \centering
    \includegraphics[width=0.75\linewidth]{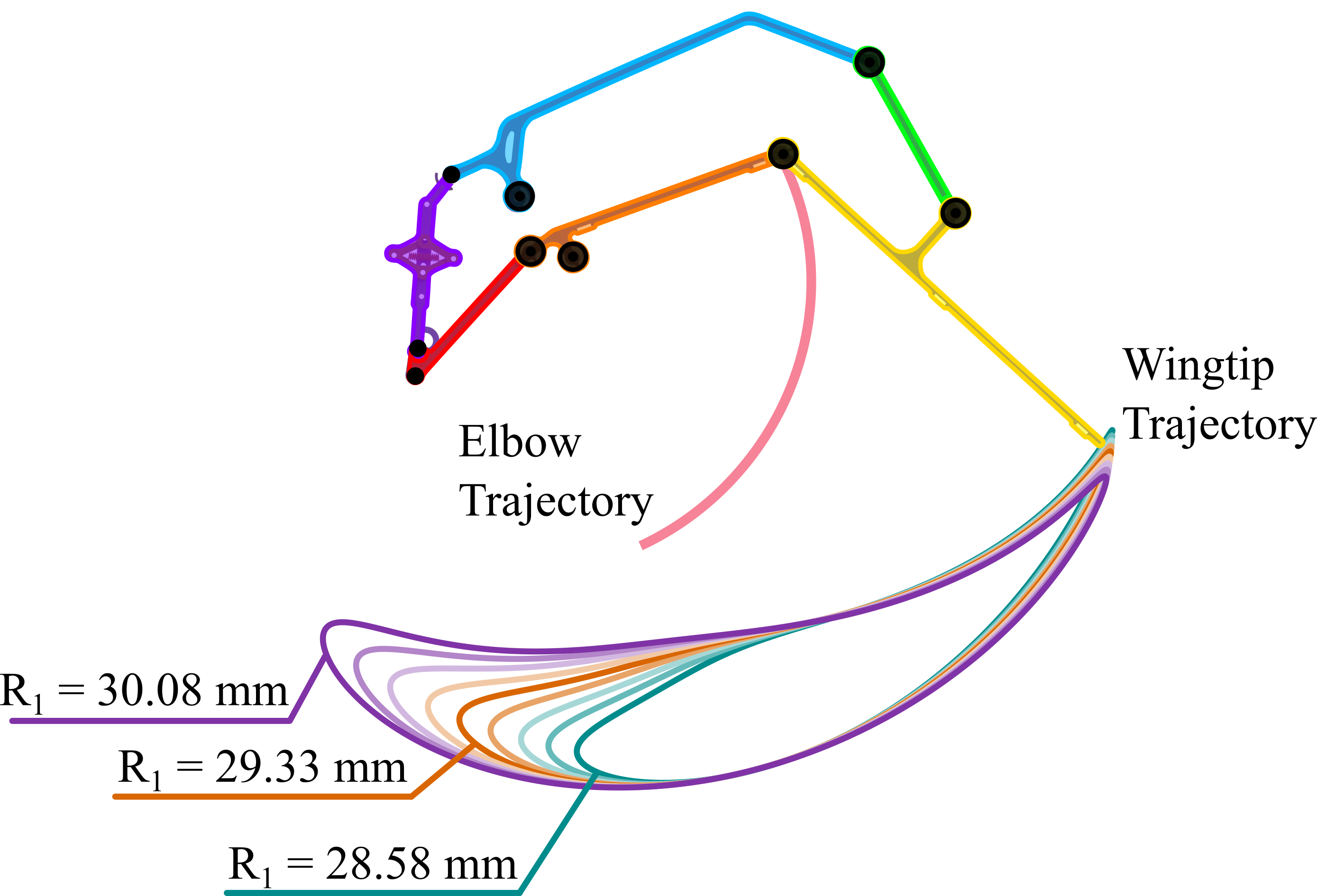}
    \caption{Effect of regulator length variation on the Aerobat flapping gait. Wingtip trajectories are shown for nine regulator link $R_1$ lengths from 28.58~mm to 30.08~mm.The computational wing structure is overlaid at a representative configuration. Larger $R_1$ lengths produce wider wingtip sweep envelopes, altering the overall stroke amplitude and flapping gait geometry.}
    \label{fig:aerobat_wing_gait_original}
\end{figure}

Changes in the length of $R_1$ alter the trajectory of the elbow joint and wingtip throughout the flapping cycle, as shown in Figure~\ref{fig:aerobat_wing_gait_original}. This radius link is driven by the motor shaft and transfers that rotary motion into the wing's kinematic chain. By replacing the original rigid $R_1$ with a variable-length prismatic joint, the effective input geometry of the mechanism can be modified during operation without altering the motor speed or any other system parameter. The lengths considered in this trajectory simulation are shown in Table~\ref{table:traj_length_matrix}. Increasing the 
length of $R_1$ expands the wingtip sweep envelope, producing a wider stroke amplitude and modifying the angle-of-attack profile throughout each flap. This results in measurable changes to both the peak aerodynamic force and the temporal distribution of force within each stroke, as confirmed by the static experiments described in Section~\ref{sec:static_testing}.

\begin{table}
\centering
\caption{Experimental Radius $R_1$ Lengths}
\label{table:icra_length_matrix}
\begin{tabular}{ |c|  }
 \hline
    Radius Length (mm) \\
 \hline
 28.58\\
 \hline
 29.33\\
 \hline
 30.08\\
 \hline
\end{tabular}
\end{table}

\begin{table}
\centering
\caption{Static Radius $R_1$ Experiment Test Matrix}
\label{table:icra_test_matrix}
\begin{tabular}{ |c|c|c|c|c|c|c|c|c|c|  }
 \hline
 Flapping Frequency &\multicolumn{3}{|c|}{3 Hz} & \multicolumn{3}{|c|}{4 Hz} &\multicolumn{3}{|c|}{5 Hz} \\
 \hline
 Test ID &\multicolumn{3}{|c|}{ Length (mm)} & \multicolumn{3}{|c|}{ Length (mm)} &\multicolumn{3}{|c|}{ Length (mm)} \\
 \hline
  Test 1 & 28.58  & 29.33  & 30.08 & 28.58  & 29.33  & 30.08 & 28.58  & 29.33  & 30.08 \\
 \hline
 
  Test 2 & 28.58  & 29.33  & 30.08 & 28.58  & 29.33  & 30.08 & 28.58  & 29.33  & 30.08 \\
 \hline
 
  Test 3 & 28.58  & 29.33  & 30.08 & 28.58  & 29.33  & 30.08 & 28.58  & 29.33  & 30.08 \\
 \hline
\end{tabular}
\end{table}

To test this directly, a subset of the radius lengths were selected to focus on a highly variable region. This selection of lengths used in the experiment is shown in Table~\ref{table:icra_length_matrix}. For each length of radius link, multiple flapping frequencies were used: 3 Hz, 4 Hz, and 5 Hz. The nominal operating frequency of the Aerobat platform is approximately 4~Hz, and the two additional frequencies were chosen to characterize the sensitivity of the link-length effect across the practical operating range. Finally, each combination of radius link length and flapping frequency was tested with 3 trials. This resulted in the experimental testing matrix shown in Table~\ref{table:icra_test_matrix}.

For the purposes of the static regulator tests, the Aerobat Delta platform was actuated with a Dynamixel XL330-M077-T servomotor. This was selected due to its relatively high output torque and rotational speed, and sensor data collecting. A gearbox was required in order to drive the input joint $J_1$ of the robot at the maximum 5 Hz, or 31.4159265 rad/s velocity. The sensors integrated into the motor provided crucial data for the experiment. This included angular position, velocity, acceleration, torque, and electric current draw. The angular velocity and current draw were critical in maintaining a steady 5 Hz flapping speed during testing.

Data collection for the experiment was handled via the 6-axis load cell and an angular encoder attached to the shoulder joint of Aerobat Delta. Measurements taken by the load cell were recorded using a desktop computer running the ATI Industrial Automation Net F/T software package. This computer was linked to the load cell using an ATI Industrial Automation Network system through an Ethernet connection. This allowed for measurements to be collected at an ideal rate of 7 kHz. The data collected was then exported as a CSV file directly from the NET F/T software package.

The shoulder angle of Aerobat Delta was recorded using a hall-effect magnetic rotary encoder. This was attached to the robot with a custom \ac{FDM}-printed mounting bracket. A magnet was fixed to the radius link, which allowed the encoder to track the relative angle of the wing. This sensor setup is shown in Fig.~\ref{fig:aerobat_on_arm_setup}. Shoulder angle measurements were recorded by an ESP-32 microcontroller connected to a computer via a USB serial link. This data was saved as a CSV file for postprocessing. In order to analyze the shoulder angle data, the maximum and minimum shoulder angle for each regulator trial had to be measured first. These measurements were kept as a calibration of the range of motion.

The load cell and encoder data acquisition systems were not hardware-synchronized during testing, as they operated on separate computers with independent clocks. Both data streams were timestamped at their respective sources and aligned during post-processing using the angular velocity zero-crossings of the shoulder joint, which correspond to the transitions between upstroke and downstroke phases. This synchronization method introduces a timing uncertainty on the order of one to two milliseconds, which at a 4~Hz flapping frequency corresponds to less than one degree of wing phase. For each experimental condition (link length and flapping frequency combination), the three trial runs were used to assess repeatability. 

\begin{figure}
    \centering
    \includegraphics[width=0.75\linewidth]{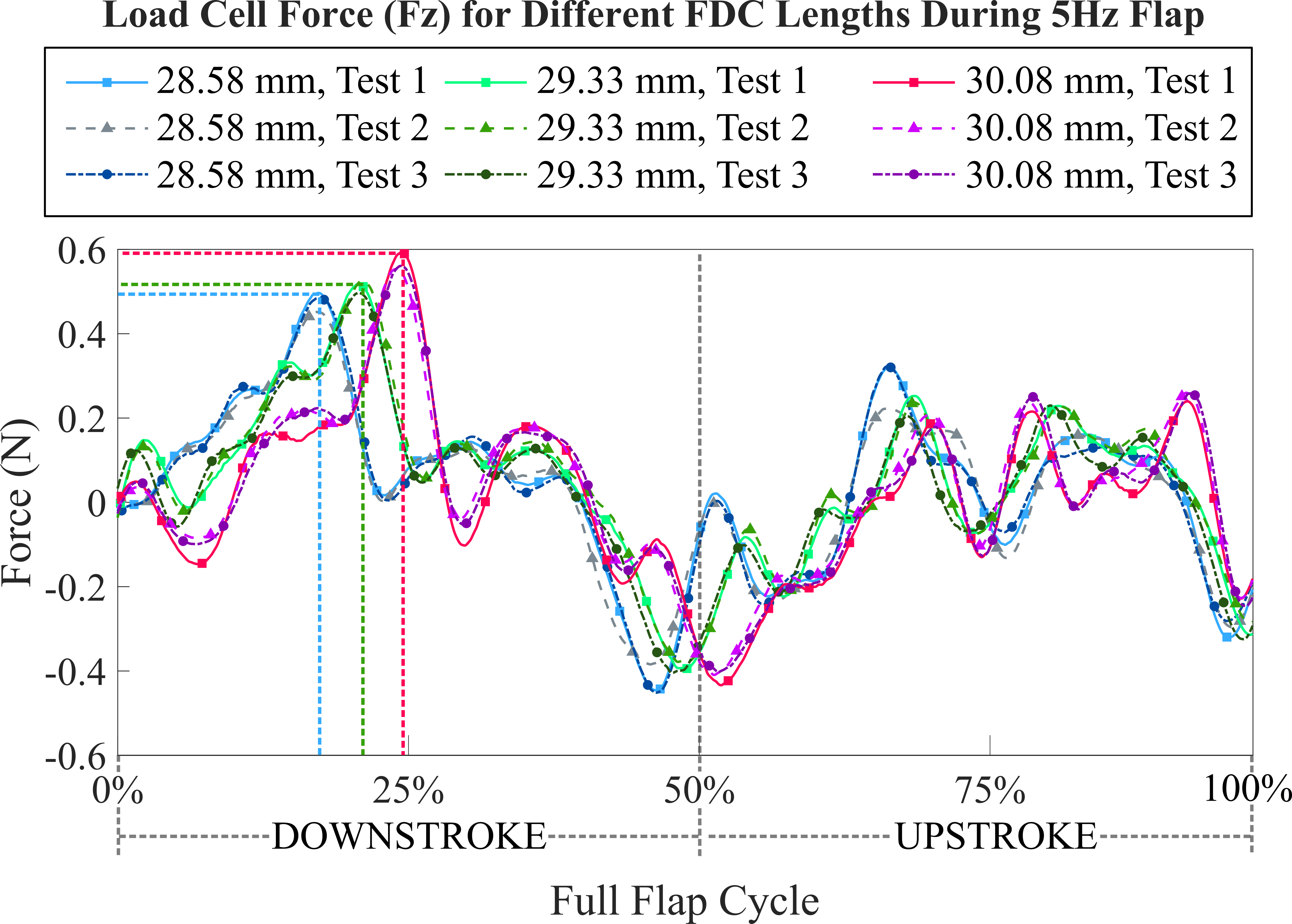}
    \caption{Static testing lift across a single 5 Hz flap cycle for all regulator lengths. The peak lift force of each regulator length is marked along with peak lift force timing.}
    \label{fig:static_test_downstroke_comparison}
\end{figure}

In order to analyze the lift produced by the Aerobat Delta prototype using fixed-length regulators, the data collected by the load cell was synchronized across the separate trial runs and lengths. This was accomplished using the shoulder joint angle data collected by the angular encoder and ESP-32 microcontroller. While this system was not synchronized with the load cell sensing during testing, both were timestamped and synchronized during post processing. The specific timings of each upstroke and downstroke were calculated based on the angular velocity of the shoulder joint. During the transitions between flap phases, the direction of the angular velocity would invert. Using this, each trial run was aligned to a selected upstroke. One upstroke and downstroke cycle of all trial runs were plotted against each other in Fig. \ref{fig:static_test_downstroke_comparison}.

Figure \ref{fig:static_test_downstroke_comparison} shows a sample of the measured vertical lift force generated during the fixed-length regulator experiments. A single flap cycle was plotted for all three trial runs of each of the three regulator lengths. The maximum regulator length of 30.08 mm produced a 37\% increase in peak force and delayed the timing of the peak from 23\% to 40\% into the downstroke. This trend was visible across all individual flaps in the collected data. 

\section{Piezoelectric Regulator Mechanism}
\begin{figure}
    \centering
    \includegraphics[width=12.5cm]{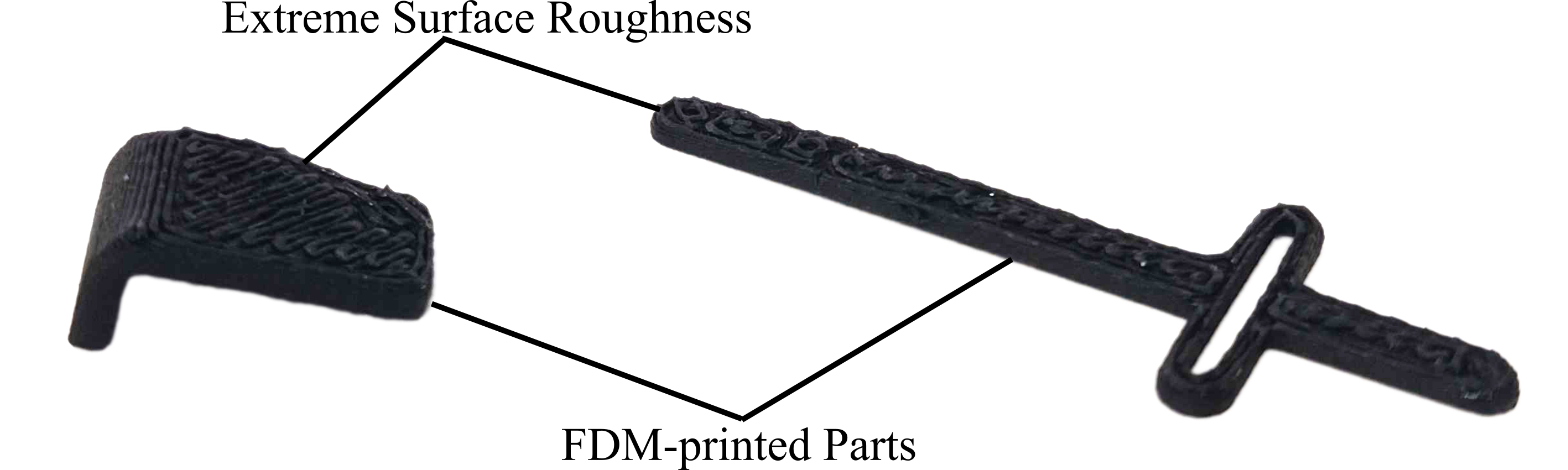}
    \caption{The extreme surface roughness in FDM-printed components due to the raft-part interface created during the printing process}
    \label{fig:fdm_part_issues}
\end{figure}

Several prototypes of both the single-triangle linkage and direct-drive piezoelectric regulator mechanisms were produced and assembled. The issues encountered during the fabrication process informed many of the decisions regarding manufacturing and component design. The first of these prototypes was a version of the single-triangle linkage fabricated using \ac{FDM} printing. This was done using a Bambu Lab X1C model of printer and the Bambu Lab proprietary carbon fiber reinforced filament. However, this prototype had a high internal friction once assembled. This was likely due to the rough surfaces of the \ac{3D}-printed components in contact with each other. This was most noticeable on areas in contact with the support material of the \ac{3D} print. Due to the small size of the components, a larger bed of printed material was required in order to provide proper build plate adhesion and prevent parts becoming loose during printing. This is referred to as a "raft". However, at the interface between each component and the raft, the surface of the parts become extremely rough and textured. This surface roughness prevented smooth operation of the first prototype. An example of this extreme surface texture is shown in Fig. \ref{fig:fdm_part_issues}.

\begin{figure}
    \centering
    \includegraphics[width=12.5cm]{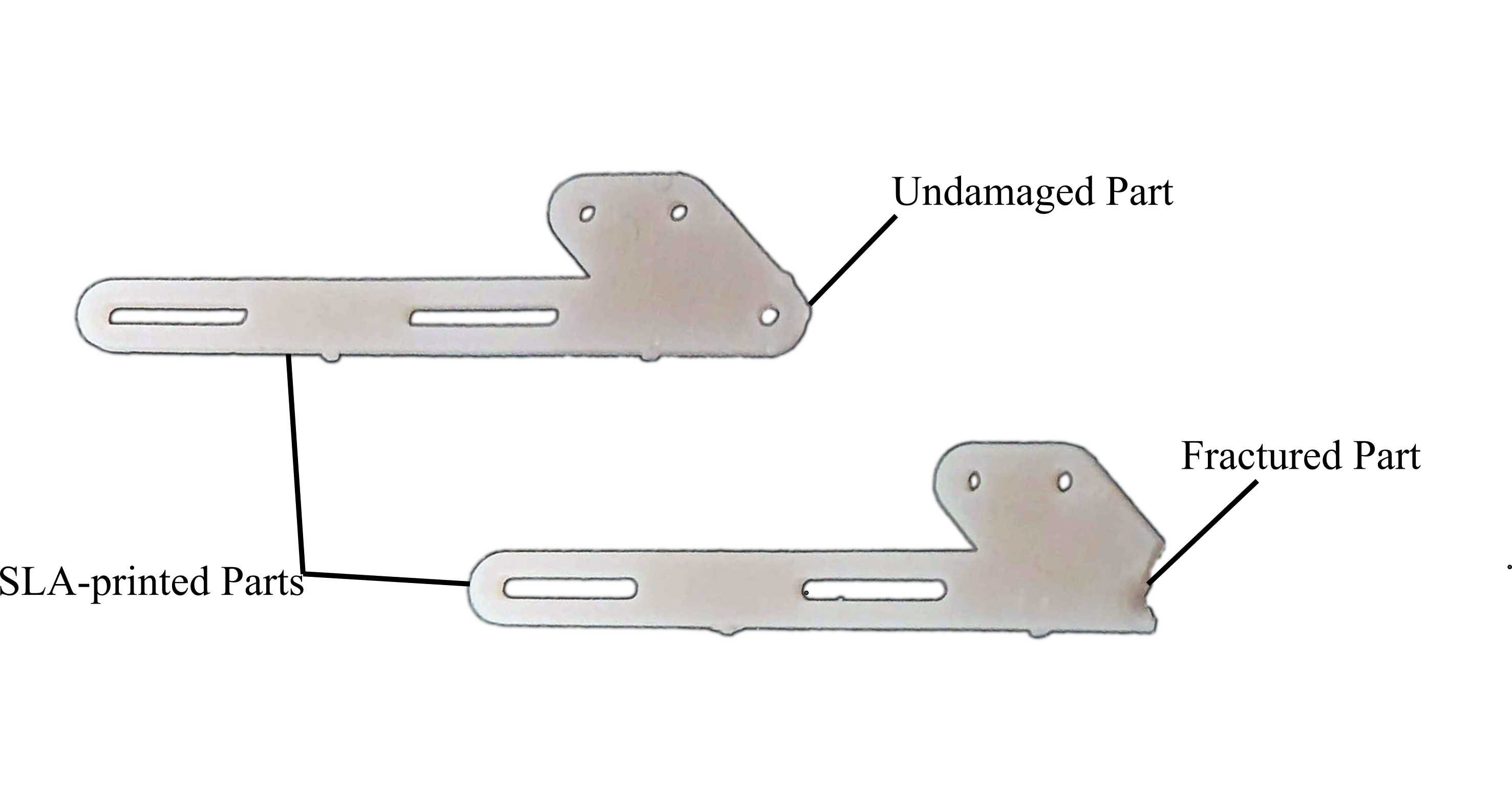}
    \caption{An undamaged and a fractured SLA-printed regulator component due to material brittleness and curing processes.}
    \label{fig:sla_part_issues}
\end{figure}

The issue of high internal friction was fixed with the second prototype of the single-triangle mechanism. This was done with \ac{SLA} printing. Unlike the \ac{FDM} printers used previously, the \ac{SLA} process results in a much lower surface friction. This the result of a combination of both the material properties of the resin used as well as the lack of a support raft. Instead, the technique uses a support scaffold that cleanly disconnects from the component once the resin is cured. While the use of this fabrication method eliminates the high friction and surface imperfections, it introduces new issues. The liquid resin that is cured to create the components becomes extremely brittle after several days of air and light exposure. This causes the thin features of the components, such as the small holes used for the steel pins during assembly, to fracture and break off. While this can be fixed at larger scales, the size limitations imposed on the regulator mechanism prevented this from becoming a viable fabrication method. Figure \ref{fig:sla_part_issues} shows an undamaged and broken copy of a SLA-printed component.

\begin{figure}
    \centering
    \includegraphics[width=0.9\linewidth]{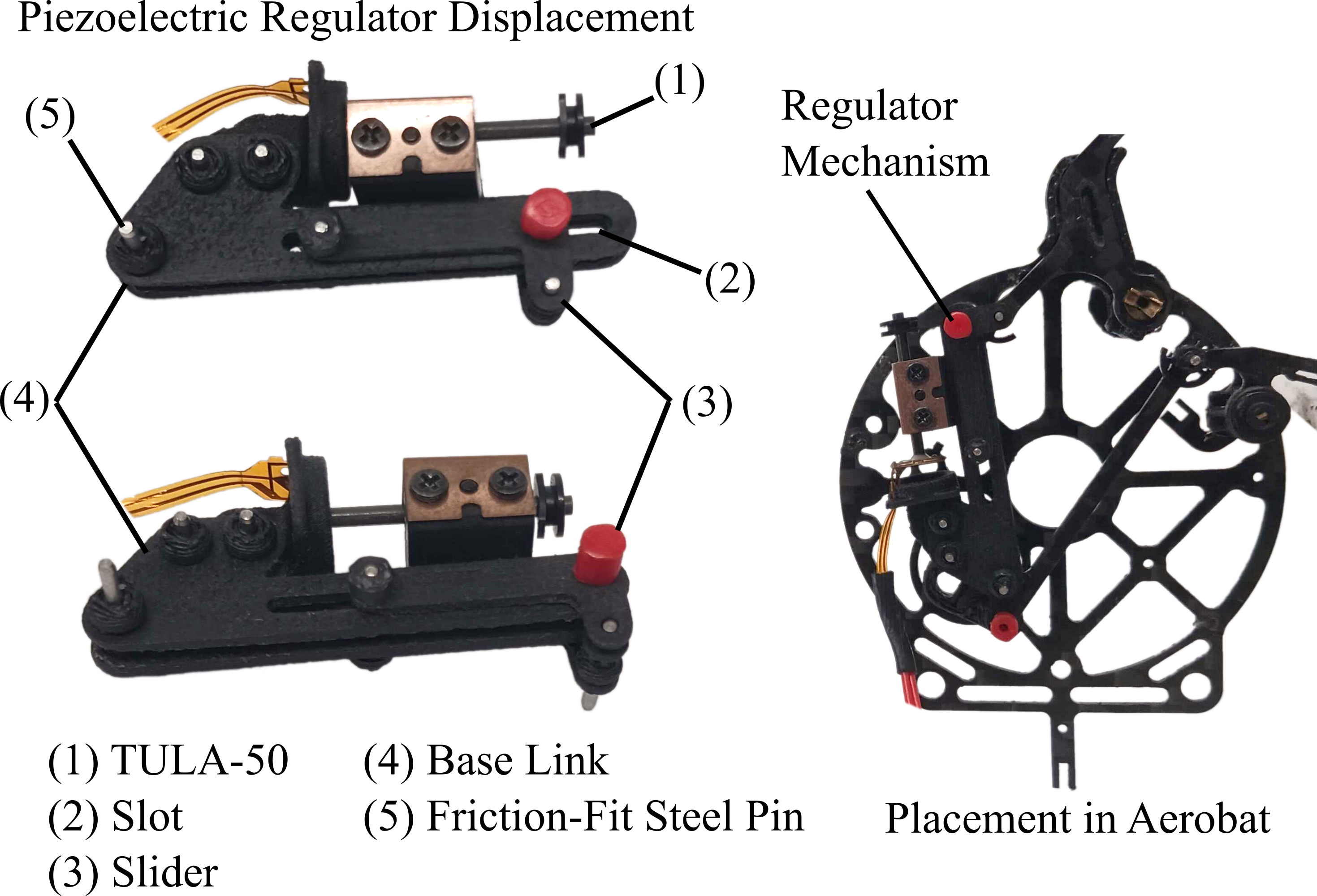}
    \caption{Final direct-drive piezoelectric slip-stick actuated regulator prototype. left top: regulator mechanism in contracted state, left bottom: regulator mechanism in extended state, right: regulator mechanism in Aerobat Delta structure}
    \label{fig:piezoelectric_regulator_real_photo}
\end{figure}

The final prototype of the piezoelectric actuator-based direct-drive type regulator mechanism is shown in Fig. \ref{fig:piezoelectric_regulator_real_photo}. For this iteration of the regulator mechanism, the primary method of fabrication used was \ac{FDM} \ac{3D} printing. This was also done using the Bambu Lab X1C printer and carbon fiber reinforced filament. The components were manually sanded smooth after printing to ensure a lower coefficient of friction and smoother motion. After processing, the components were secured with steel pins fabricated from thin 1 mm diameter wire. As sown in Fig.~\ref{fig:piezoelectric_regulator_real_photo}, these are held in place by a friction fit wit the \ac{3D}-printed slider and base links. Small \ac{3D}-printed retaining rings were also used to secure the components in place. These also acted as spacers for the points of contact with the other links in Aerobat Delta.

This mechanism was tested outside the computational structure of Aerobat Delta. Without the constraints of the linkage structure, it was able to actuate itself and dynamically change length over the whole range of motion. However, the actuator had difficulty in covering the entire range of motion once integrated into the larger kinematic structure. An oscillating sinusoidal command was given to the \ac{TULA}-50 actuator to test the range of motion. While placed in the computational structure, the regulator mechanism was unable to consistently cover the full range of motion in a smooth, uninterrupted movement. This is due to the larger range of motion than intended for the regulator mechanism as a result of the direct-drive approach. This also reduced the mechanical advantage of the mechanism from between 4:1 and 14:1 down to 1:1, making it unable to function while the wing was in motion as it could no longer produce the required 295.89~gf. Figure \ref{fig:piezoelectric_regulator_real_photo} shows the prototype integrated into the Aerobat Delta test platform. Due to both time and material constraints, as well as the weakness of the overall mechanism after the design revisions, this was the extent of testing performed on the final design. As demonstrated in the figure, the extension of the piezoelectric actuator drives the motion of the $R_2$ radius link about the $J_5$ shoulder joint. This motion flexes the wing structure at humerus joint $J_8$. 

% conclusion and Future work
 \chapter{Conclusion}
\label{chap:conclusion}

Based on the results of the tests done for this thesis, the underlying problem across all proposed solutions is actuator reliability. 

While the initial experiment with varying radius lengths served as proof that the thrust could be regulated in this manner, further testing with the completed mechanism prototypes was not possible. This was primarily due to the various issues with the existing Aerobat Delta test platform and fabrication methods available for this project. 

The primary issue found with Aerobat Delta was the structural integrity of the robot. Due to issues with the kinematic structure, joints consistently failed during flapping. This also led to the axles and pulleys used in the tension-actuated prototype to tilt out-of-plane in their sockets, negatively affecting the tension regulation in the robot. The flexibility in both the carbon fiber links and revolute joints also led to out-of-plane rotation of the wing panels due to uneven aerodynamic loading. In turn, this induced significant vibrations in the rest of the linkage components that regularly caused the string to disengage from the pulleys during flapping. The out-of-plane forces and moments that were observed acting on the regulator mechanism also influenced the selection of later actuation methods, as any mechanism had to be sufficiently resistant to this out-of-plane loading. The lack of resilience was one of the primary concerns with the sub-gram micro-servo prototype. The fragility of the motor frame was a limiting factor in the performance of the mechanism, and led to the discontinuation of their testing. 

There are several avenues for related future work. Adjacent to the scope of this thesis, validating the performance of several more actuators would potentially yield positive results. While the the piezoelectric actuation method showed promise, the \ac{TULA}-50 actuator model that was tested had a relatively low force yield of 20~grams-force. Selecting a larger \ac{TULA} model would provide more actuation force while having minimal impact on the size or mass of the overall mechanism, as the \ac{TULA}-70 model has a higher maximum load of 50~grams-force. This would increase both the force produced by the actuator in addition to the stroke length with a minimal increase in mass and complexity. Another potential extension of this research is exploring different fabrication methods. Utilizing different manufacturing techniques would also address several of the issues encountered in this thesis, as the use of \ac{3D}-printing limited the material selection and geometrical resolution. Fabricating the regulator components from the same carbon fiber composite as the rest of the structure would allow for the mechanism to be both smaller and more robust in operation. It would also eliminate issues surrounding the flexibility of \ac{SLA} resin components.

The future of this research also includes the integration of the regulator mechanism into the structure of a two-winged Aerobat platform. The usage of multiple regulators on a robot would allow for a much greater degree of control over the orientation during flight. In addition, regulator mechanisms could be integrated into other segments of the computational structure to alter the gait even further. The combination of multiple regulators on each wing could therefor potentially allow for control over multiple degrees of freedom in the air.

% --- Bibliography ----
\bibliographystyle{IEEEtran}  %'plain' for standard, 'unsrt' for correct order

% include bibliography definition
\bibliography{bib/thesis}

% --- Appendix ---
\appendix
%include anything you need in the appendix
%\include{tex/appendixA}

% --- Index ----
%\printindex

% --- that's it ---
\end{document}